\documentclass[Proceedings]{ascelike}
\usepackage{setspace}
\usepackage[displaymath, mathlines]{lineno}
\linenumbers
\usepackage{amssymb, bm, amsmath, lipsum}
\usepackage[mathscr]{euscript}
\usepackage{subcaption}
\renewcommand{\vec}[1]{\boldsymbol{\mathbf{#1}}} 

\usepackage{tikz}

\usepackage{graphicx}
\usetikzlibrary{shapes,arrows}
\usetikzlibrary{fit,positioning}
\usetikzlibrary{calc,patterns,decorations.pathmorphing,decorations.markings}

\usepackage{makecell}
\usepackage{algpseudocode}
\usepackage[vlined, linesnumbered]{algorithm2e}
\makeatletter
\renewcommand{\@algocf@capt@plain}{above}
\makeatother

\usepackage{xparse}
\DeclareDocumentCommand{\gaussianDist}{m m g}{ 
    {\mathcal{N}\left(\IfNoValueF{#3}{#3 \, \vert \,}
        #1,#2\right)%
    }
}

\usepackage[hidelinks=true]{hyperref}
\nolinenumbers

\NameTag{Bull, \today}
\begin{document}

\DeclareDocumentCommand{\gaussianDist}{m m g}{ 
	{\mathcal{N}\left(\IfNoValueF{#3}{#3 \, \vert \,}
		#1,#2\right)%
	}
}
\newcommand{\gammaDist}[2]{\mathcal{G}\left(#1,#2\right)} 

\usetikzlibrary{arrows.meta,
	bending,
	calc, chains,
	decorations.pathmorphing,  
	positioning
}
\tikzstyle{X1} = [rectangle, minimum width=2.3cm, minimum height=1.2cm,text centered, align=center,fill=gray!20]
\tikzstyle{XT} = [rectangle, minimum width=2cm, minimum height=1cm,text centered, align=center,fill=gray!20]
\tikzstyle{K1} = [rectangle, minimum width=1.2cm, minimum height=1.2cm,text centered, align=center,fill=gray!20]
\tikzstyle{KT} = [rectangle, minimum width=1cm, minimum height=1cm,text centered, align=center,fill=gray!20]
\definecolor{c1}{rgb}{0.9843,    0.7059,    0.6824}
\tikzstyle{A1} = [rectangle, minimum width=1.2cm, minimum height=0.8cm,text centered, align=center,fill=c1]
\tikzstyle{AT} = [rectangle, minimum width=1cm, minimum height=0.8cm,text centered, align=center,fill=c1]
\tikzstyle{B} = [rectangle, minimum width=0.5cm, minimum height=0.5cm,text centered, align=center,fill=c1]
\tikzstyle{W} = [rectangle, minimum width=0.5cm, minimum height=0.8cm,text centered, align=center,fill=c1]
\definecolor{c2}{rgb}{0.7020,    0.8039,    0.8902}
\tikzstyle{H1} = [rectangle, minimum width=0.8cm, minimum height=1.2cm,text centered, align=center,fill=c2]
\tikzstyle{HT} = [rectangle, minimum width=0.8cm, minimum height=1cm,text centered, align=center,fill=c2]
\definecolor{c2}{rgb}{0.7020,    0.8039,    0.8902}
\tikzstyle{f1} = [rectangle, minimum width=0.5cm, minimum height=1.2cm,text centered, align=center,fill=c2]
\tikzstyle{fT} = [rectangle, minimum width=0.5cm, minimum height=1cm,text centered, align=center,fill=c2]
\tikzstyle{y1} = [rectangle, minimum width=0.5cm, minimum height=1.2cm,text centered, align=center,fill=gray!20]
\tikzstyle{yT} = [rectangle, minimum width=0.5cm, minimum height=1cm,text centered, align=center,fill=gray!20]
%
\title{PROBABILISTIC INFERENCE FOR STRUCTURAL HEALTH MONITORING: NEW MODES OF LEARNING FROM DATA}
\author{
Lawrence A.\ Bull, Paul Gardner, Timothy J.\ Rogers \\
Elizabeth J.\ Cross, Nikolaos Dervilis, Keith Worden}

\maketitle

{\scriptsize Dept.\ of Mech.\ Eng., Univ.\ of Sheffield, Mappin St., Sheffield, S1 3JD, UK 
\href{mailto:l.a.bull@sheffield.ac.uk}{l.a.bull@sheffield.ac.uk}}

\vspace{1em}
\begin{center}
{\scriptsize This material may be downloaded for personal use only. Any other use requires prior permission of the American Society of Civil Engineers. This material may be found at \url{https://doi.org/10.1061/AJRUA6.0001106}}
\end{center}

\begin{abstract}
    In data-driven SHM, the signals recorded from systems in operation can be noisy and incomplete. Data corresponding to each of the operational, environmental, and damage states are rarely available \textit{a priori}; %
    furthermore, labelling to describe the measurements is often unavailable.
    In consequence, the algorithms used to implement SHM should be robust and adaptive, while accommodating for missing information in the training-data -- such that new information can be included if it becomes available. %
    By reviewing novel techniques for statistical learning %
    (introduced in previous work), it is argued that probabilistic algorithms offer a natural solution to the modelling of SHM data in practice. %
    In three case-studies, probabilistic methods are adapted for applications to SHM signals --- including semi-supervised learning, active learning, and multi-task learning.
\end{abstract}

\vspace{1em}
\section{Probabilistic SHM}
Under the pattern recognition paradigm associated with Structural Health Monitoring (SHM) \cite{SHM}, data-driven methods have been established as a primary focus of research. Various machine learning tools have been applied in the literature, for example \cite{vanik2000bayesian,sohn2003review,chatzi2009unscented}, and used to infer the health or performance state of the monitored system, either directly or indirectly. Generally, algorithms for regression, classification, density estimation, or clustering learn patterns in the measured signals (available for training), and the associated patterns can be used to infer the state of the system in operation, given future measurements \cite{worden2006application}.

Unsurprisingly, there are numerous ways to apply machine learning to SHM. Notably (and categorised \textit{generally}), advances have focussed on various probabilistic (e.g.\ \cite{vanik2000bayesian,ou2017vibration,flynn2010bayesian}) and deterministic (e.g.\ \cite{bornn2009structural,zhao2019deep,janssens2017deep}) methods. %
Each approach has its advantages; however, considering certain challenges associated with SHM data (outlined in the next section) the current work focusses on probabilistic (i.e.\ statistical) tools: these algorithms appear to offer a natural solution to some key issues, which can otherwise prevent practical implementation. %
Additionally, probabilistic methods can lead to predictions \textit{under uncertainty} \cite{papoulis1965} -- a significant advantage in risk-based applications. %

\subsection{SHM, Uncertainty, and Risk}
It should be clear that measured/observed data in SHM will be inherently uncertain, to some degree. Uncertainties can enter via \textit{experimental} sources, including limitations to sensor accuracy, precision or human error; further uncertainties will be associated with the model -- machine learning or otherwise -- including parametric variability, model discrepancy, and interpolation uncertainty. %
Considering the implications of \textit{risk}, financially and in terms of safety, uncertainty should be mitigated (during data acquisition), and quantified (within models) as far as possible to inform decision making \cite{zonta2014value,cappello2015mechanical}. %
That is, when supporting a financial or safety-critical decision, predictions should be presented with \textit{confidence}: clearly, a certain prediction, which implies a system is safe to use, differs significantly to an \textit{uncertain} prediction, supporting the same decision. If there is no attempt to quantify the associated uncertainties, there is no distinction between these scenarios. %

Various methods can return predictions with confidence (or \textit{credibility}) \cite{murphy}. The current work focusses on probabilistic models, which -- under Kolmogorov's axioms \cite{papoulis1965} -- allow for predictions under well-defined uncertainty, provided the model assumptions are \textit{appropriate}.

\subsection{A Probabilistic Approach}
\begin{figure}[pt]
    \centering
    \resizebox{\textwidth}{!}{%
    \begin{tikzpicture}[auto]
        \linespread{1}
    \tikzstyle{block} = [rectangle, thick, draw, text width=6em, text centered, minimum height=7em]
    \tikzstyle{block3} = [rectangle, thick, draw=black!40, text width=6em, text centered, minimum height=6em]
    \tikzstyle{block2} = [draw=black!0, rectangle, text width=5em, text centered, minimum height=6em]
    \tikzstyle{line} = [draw, -latex, thick]
    
        \node [block] (SHM) { $f(\cdot)$ \\ pattern recognition};
        \node [block3, left of=SHM, node distance=45mm] (SP) {pre-processing, feature extraction};
        \node [block2, left of=SP, node distance=30mm] (MD) {measured data};
        \node [block3, right of=SHM, node distance=45mm] (PP) {post-\\processing};
        \node [block2, right of=PP, node distance=30mm] (label) {diagnostic labels};
        \path [line] (SP) -- node [above] {\textsl{inputs}} node [below] {$\vec{x}_i$} (SHM);
        \path [line, draw=black!40] (MD) -- (SP);
        \path [line, draw=black!40] (PP) -- (label);
        \path [line] (SHM) --  node [above] {\textsl{outputs}} node [below] {$y_i$} (PP);
    
    \end{tikzpicture}
    }%
    \caption{A \textsl{simplified} framework for pattern recognition within SHM.}
    \label{SHM_khart}
\end{figure}
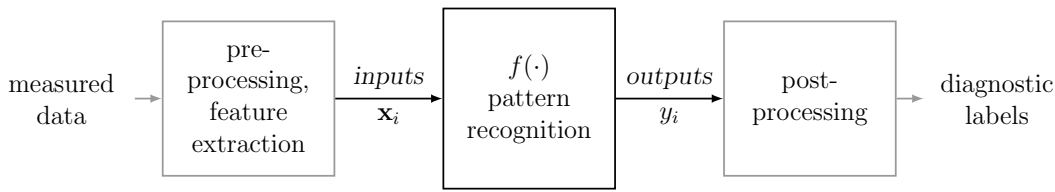

Discussions in this work will consider the general strategy illustrated in Figure~\ref{SHM_khart}. That is, SHM is viewed as a multi-class problem, which categorises measured data into groups, corresponding to the condition of the monitored system. %
The $i^{th}$ input, denoted by $\vec{x}_i$, is defined by a $d$-dimensional vector of variables, which represents an \textit{observation} of the system, such that $\vec{x}_i \in \mathbb{R}^d$. %
The data \textit{labels} $y_i$, are used to specify the condition of the system, directly or indirectly. %
Machine learning is introduced via the pattern recognition model, denoted $f(\cdot)$, and is used to infer relationships between the input and output variables, to inform predictive maintenance.

The inputs $\vec{x}_i$ are assumed to be represented by some random vector $X$ (in this case, a continuous random vector), which can take any value within a given feature-space $\mathscr{X}$. The random vector is therefore associated with an appropriate probability density function (p.d.f.), denoted $p(\cdot)$, %
such that the probability $P$ of $X$ falling within the interval $a < X \leq b$ is, %
$P\left(a < X \leq b \right)\; = \int_{a}^{b} p\left(\vec{x}_i\right)\,d\vec{x}_i \;\textrm{such that}\; p\left(\vec{x}_i\right)\geq 0,\; \int_{_\mathscr{X}} p\left(\vec{x}_i\right)\,d\vec{x}_i = 1$. %
For a discrete classification problem, the labels $y_i$ are represented by a discrete random variable $Y$, which can take any value from the finite set, $y_i \in \mathscr{Y} = \{1,...,{K}\}$. Note: {discrete classification is presented in this work, although, SHM is regularly informed by regression models -- i.e.\ $y_i$ is continuous; this is application specific, and most of the motivational arguments remain the same.} $K$ is the number of classes defining the (observed) operational, environmental, and health conditions, while $\mathscr{Y}$ denotes the label-space. An appropriate probability mass function (p.m.f.), also denoted $p(.)$, is such that, %
$P\left({Y} = y_i\right) = p(y_i) \;\textrm{where}\; 0\leq P\left({Y} = y_i\right)\leq 1,\; \sum_{y_i\in Y} P\left({Y} = y_i\right) = 1$. %
Note: {context should make the distinction between p.m.fs and p.d.fs clear.} %
Further details regarding probability theory for pattern recognition can be found in a number of well written textbooks -- for example \cite{murphy,barber2012bayesian,gelman2013bayesian}. %

\subsection{Layout}
Section~\ref{s:sparse_data} summarises the most significant challenges for data-driven SHM, while Section~\ref{s:intro} suggests probabilistic methods to mitigate these issues. %
Section~\ref{s:DGM} introduces theory behind directed graphical models (DGMs), which will be used to formally introduce each method. %
Section~\ref{s:case_studies} collects four case studies to highlight the advantages of probabilistic inference. Active learning and Dirichlet process clustering are applied to the Z24 bridge data. %
Semi-supervised learning is applied to data recorded during ground vibration tests of a Gnat aircraft. %
Multi-task learning is applied simulated and experimental data from shear-building structures. %

Note: the applications presented here were introduced in previous work by the authors. The related SHM literature is referenced in the descriptions of each mode of inference. %

\section{Incomplete Data and Missing Information}\label{s:sparse_data}
Arguably, the most significant challenge when implementing pattern recognition for SHM is missing information. %
Primarily, it is difficult to collect data that might represent {damage states} or the system in extreme {environments} (such as earthquakes) \textit{a priori}; %
data are usually only available for a limited subset of the possible conditions for training algorithms \cite{SHM}. %
As a result, conventional methods are restricted to novelty detection, %
as the information required to inform \textit{multi-class} predictive models (that can localise and classify damage, as well as detect it \cite{worden2006application}) is unavailable or not obtained.

For the measurements $\vec{x}_i$ that are available -- as well as those that are recorded during operation (\textit{in situ}) -- \textit{labels} to describe what the signals represent, $y_i$, are rarely at hand. %
This missing information is usually due to the cost associated with manually inspecting structures (or data), as well as the practicality of investigating each observation. %
The absence of labels makes defining and updating (multi-class) machine learning models difficult, particularly in the online setting, %
as it can become difficult to determine if/when novel valuable information has been recorded, and what it represents \cite{onlineAL}. %
For example, consider streaming data, recorded from a sub-sea pipeline. Comparisons of measured data to the model might indicate novelty; however, without labels, it is difficult to include this new information in a supervised manner: the measurements might represent another operational condition, abnormal wave loads, actual damage, or some other condition.

\section{New Modes of Probabilistic Inference}\label{s:intro}
New modes of probabilistic inference are being proposed to address challenges with SHM data. %
Specifically, the algorithms focus on probabilistic frameworks to deal with \textit{limited labelled data}, as well as \textit{incomplete measured data}, that only correspond to a subset of the expected conditions \textit{in situ}.

\subsection{Partially-Supervised Learning}
\textit{Partially-supervised learning} allows multi-class inference in cases where labelled data are limited. %
Missing label information is especially relevant to practical applications of SHM: while \textit{fully} labelled data are often infeasible, it can be possible to include labels for a limited set (or \textit{budget}) of measurements. %
Typically, the budget is limited by some expense incurred when investigating the signals; this might include direct costs associated with inspection, or loss of income due to down-time \cite{BULL2020106653}. 

Generally speaking, partially-supervised methods can be used to perform multi-class classification, while utilising \textit{both} labelled $\mathcal{D}_l$ and unlabelled $\mathcal{D}_u$ signals within a \textit{unifying} training scheme \cite{Schwenker2014} -- as such, the training set $\mathcal{D}$ becomes,

\begin{align}
    \mathcal{D} &=\mathcal{D}_l \cup \mathcal{D}_u \\
    &= \left\{\vec{X},\vec{y}\right\} \cup \tilde{\vec{X}}\\[1em]
    \left\{\vec{X},\vec{y}\right\} &\triangleq \left\{\vec{x}_i,y_i\right\}_{i=1}^{n}\\
    \tilde{\vec{X}} & \triangleq \left\{\tilde{\vec{x}}_i \right\}_{i=1}^{m} 
\end{align}

\textit{Active} and \textit{semi-supervised} techniques are suggested -- as two variants of partially-supervised learning -- to combine/include information from labelled and unlabelled SHM data \cite{bull2018active,onlineAL,BULL2020106653}.

\subsubsection{Semi-supervised learning}
Semi-supervised learning utilises \textit{both} the labelled and unlabelled data to inform a classification \textit{mapping}, $f: \mathscr{X} \mapsto \mathscr{Y}$. %
Often, a semi-supervised learner will use information in $\mathcal{D}_u$ to further update/constrain a classifier learnt from $\mathcal{D}_l$ \cite{mccallumzy1998employing}, or, alternatively, partial supervision can be implemented as constraints on a \textit{unsupervised} clustering algorithm \cite{SS}. This work focusses on classifier-based methods; however, constraints on clustering algorithms are discussed in later sections.

Arguably, the most simple/intuitive method to introduce unlabelled data is \textit{self-labelling} \cite{zhu2005semi}. In this case, a classifier is trained using $\mathcal{D}_l$, which is used to predict labels for the unlabelled set $\mathcal{D}_u$. %
This defines a new training-set -- some labels in $\mathcal{D}$ are the ground truth, from the supervised data, and the others are \textit{pseudo-labels}, predicted by the classifier. %
Self-labelling is simple, and it can be applied to any supervised method; however, the effectiveness is highly dependent on the method of implementation, and the supervised algorithm within it \cite{SS}.

\begin{figure}[pt]
    \centering
    \begin{subfigure}[b]{.49\textwidth}
      \centering
      \includegraphics[width=\linewidth]{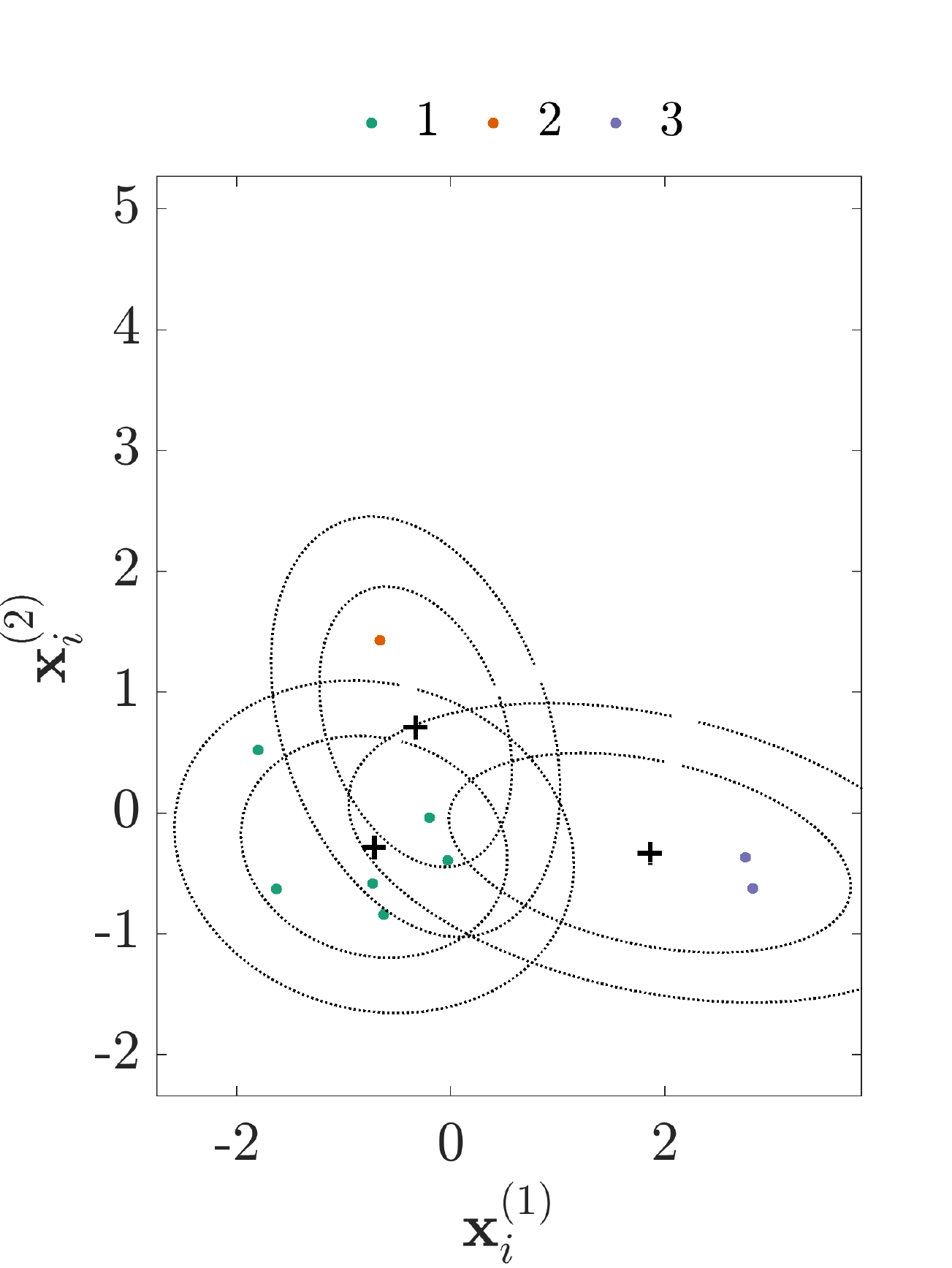}
      \caption{\label{a}}\label{fig:gmm_sl}
    \end{subfigure}
    \begin{subfigure}[b]{.49\textwidth}
      \centering
      \includegraphics[width=\linewidth]{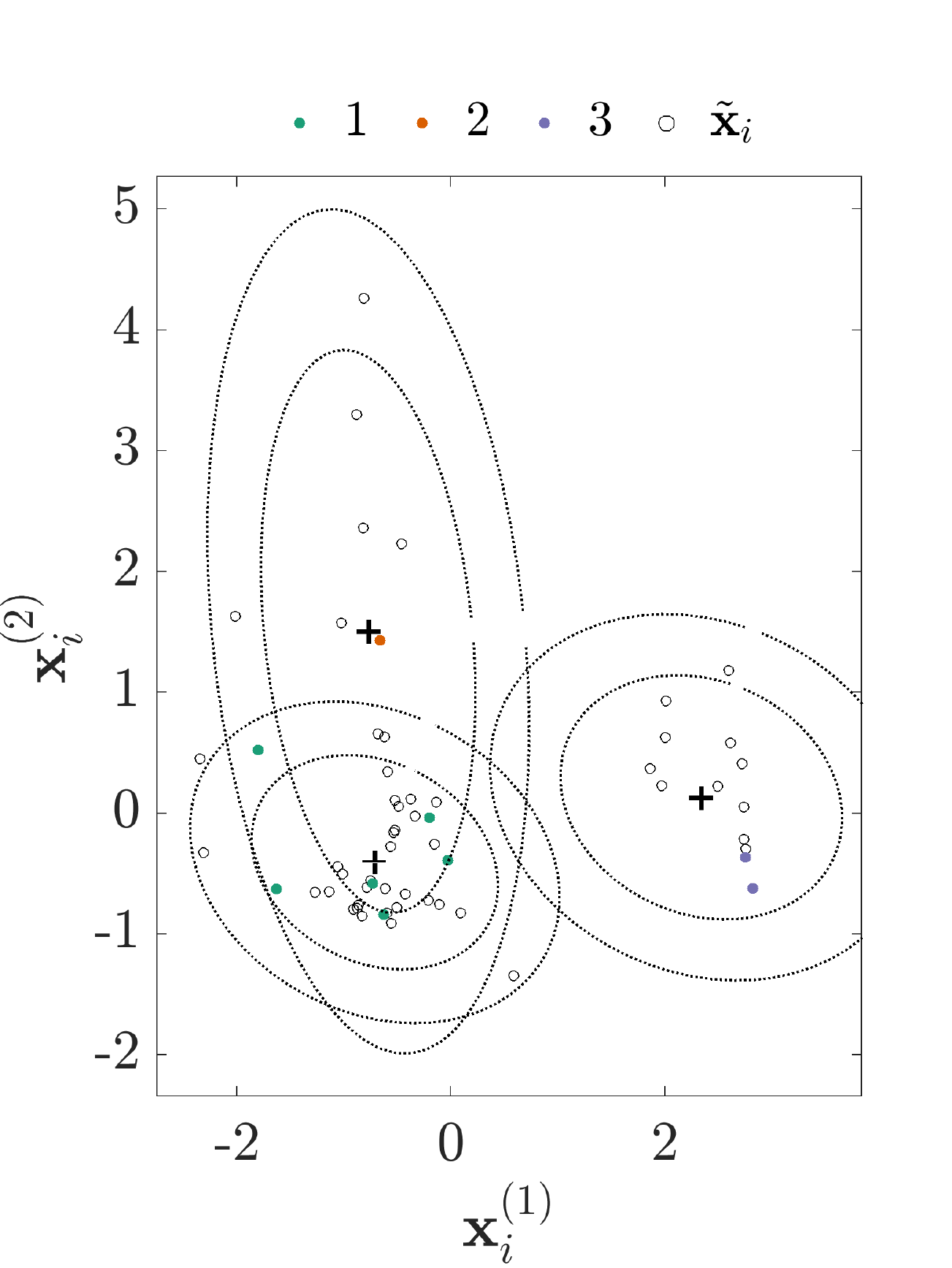}
      \caption{\label{b}}\label{fig:gmm_ssl}
    \end{subfigure}
    \caption{Semi-supervised GMM for three-class AE data: (\subref{a}) supervised learning, given the labelled data only, $\bullet$ markers. (\subref{b}) semi-supervised learning, given the labelled \textsl{and} unlabelled data, $\bullet/ \circ$ markers. Adapted from \protect\cite{bull_2019thesis}.}\label{fig:gmm_ss_eg}
\end{figure}

Generative mixture models offer a formal \textit{probabilistic} framework to incorporate unlabelled data \cite{cozman2003semi,nigam1998learning}. %
Generative mixtures apply the \text{cluster assumption}: \textsl{`if points are in the same cluster, they are likely to be of the same class}'. %
Note: {the cluster assumption does not necessarily imply that each class is represented by a single, compact cluster; instead, the implication is that observations from different classes are unlikely to appear in the same cluster \cite{SS}.}
Through density estimation \cite{barber2012bayesian}, a mixture of base-distributions can be used to estimate the underlying distribution of the data, $p(\vec{x}_i, y_i)$, and unlabelled observations can be included in various ways \cite{mccallumzy1998employing,vlachos2009unsupervised}.  
For example, the Expectation Maximisation (EM) algorithm (used to learn mixture models in the unsupervised case \cite{murphy}) can be modified to incorporate labelled observations \cite{nigam1998learning,mccallumzy1998employing}. %
Figure~\ref{fig:gmm_ss_eg} demonstrates how a Gaussian mixture, given acoustic emission data \cite{AE}, can be improved by considering the surrounding unlabelled examples (via EM). %

To summarise, semi-supervised methods allow algorithms to learn from information in the available unlabelled measurements as well as a limited set of labelled data. %
In practice, semi-supervised inference implies that the cost associated with labelling data could be managed in SHM \cite{chen2013,chen2014}, as the information in a small set of labelled signals is combined with larger sets of unlabelled data \cite{bull2019damage}.

\subsubsection{Active Learning}
\begin{figure}[pt]
    \centering
    \resizebox{\textwidth}{!}{%
    \begin{tikzpicture}[auto]
    \tikzstyle{block} = [rectangle, thick, draw=black!80, text width=8em, text centered, minimum height=7em, fill=black!5]
    \tikzstyle{line} = [draw, -latex, thick]
    
        \node [block, node distance=42mm] (A) {provide\\ unlabelled input data};
        \node [block, right of=A, node distance=42mm] (B) {establish which data are the most informative};
        \node [block, right of=B, node distance=42mm] (C) {provide labels for these data};
        \node [block, right of=C, node distance=42mm] (D) {train a classifier on this informed subset};
        \path [line, draw=black!80] (A) -- (B);
        \path [line, draw=black!80] (B) -- (C);
        \path [line, draw=black!80] (C) -- (D);
        \path [line, draw=black!80, dashed]  (D) -- node {} ++(0,-2cm) -| (A) node[pos=0.25] {} node[pos=0.75] {};
    
    \end{tikzpicture}
    }
    \caption{A general/simplified active learning heuristic.}
    \label{AL_frmwrk}
\end{figure}
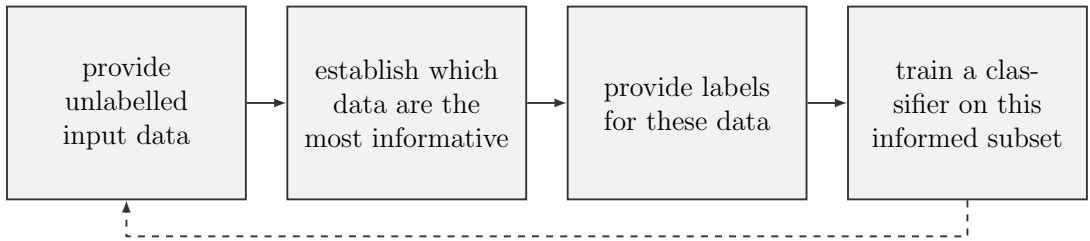

Active learning is an alternative partially-supervised method; the key hypothesis is that an algorithm can provide improved performance, using fewer training labels, if it is allowed to select the data from which it learns \cite{settles2012active}. %
As with semi-supervised techniques, the learner utilises $\mathcal{D}_l$ and $\mathcal{D}_u$ -- however, active algorithms query/annotate the unlabelled data in $\mathcal{D}_u$ to extend the labelled set $\mathcal{D}_l$. Thus, an active learner attempts to define an accurate mapping, $f: \mathscr{X} \mapsto \mathscr{Y}$, while keeping queries to a minimum \cite{two_faces}; general (and simplified) steps are illustrated in Figure~\ref{AL_frmwrk}. %

\begin{figure}[pt]
	\centering
	\centering
  \begin{subfigure}{.45\textwidth}
  \centering
  \includegraphics[width=\linewidth]{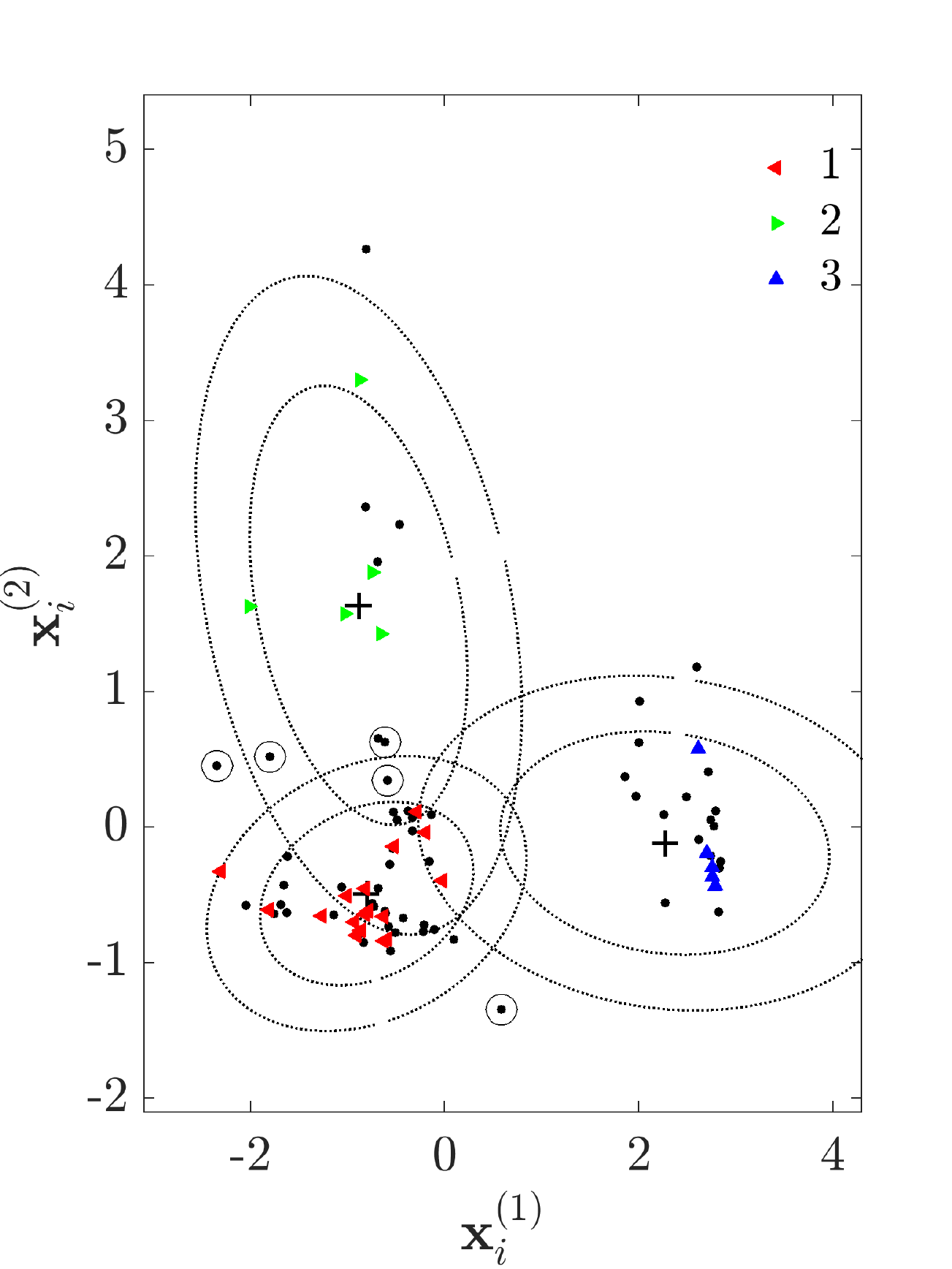}
  \caption{\label{a}}
  \label{fig:entQ}
  \end{subfigure}
  \begin{subfigure}{.45\textwidth}
  \centering
  \includegraphics[width=\linewidth]{figures/fig_4a.pdf}
  \caption{\label{b}}
  \label{fig:likQ}
  \end{subfigure}
	\caption{Uncertainty sampling for the AE data: {$\blacktriangleright\;\blacktriangleleft \blacktriangledown$~markers} show the training set, and $\bullet$ markers show the unlabelled data -- circles indicate queries by the active learner (a) based on entropy, (b) based on likelihood -- adapted from \protect\cite{bull_2019thesis}.}\label{fig:al_ae}
\end{figure}

The critical step for active algorithms is how to select the most informative signals to investigate \cite{wang_density,Schwenker2014}. 
For example, \textit{Query by Committee (QBC)} methods build an ensemble/committee of classifiers using a small, initial (random) sample of labelled data, leading to multiple predictions for unlabelled instances. Observations with the most conflicted label predictions are viewed as informative, thus, they are queried \cite{wang_density}. %
On the other hand, \textit{uncertainty-sampling} usually refers to a framework that is based around a single classifier \cite{kremer_asvm,settles2012active}, where signals with the \textit{least confident} predicted label, given the model, are queried. %
(It is acknowledged that QBC methods can also be viewed as a type of uncertainty sampling.) %
Uncertainty sampling is (perhaps) most interpretable when considering probabilistic algorithms, as the posterior probability over the class-labels $p(y_i\,|\,\vec{x}_i)$ can be used to quantify uncertainty/confidence  \cite{bull2020investigating}. %
For example, consider a binary (two-class) problem: intuitively, uncertain samples could be instances whose posterior probability is nearest to $0.5$ for both classes. This view can be extended to multiple ($> 2$) classes using the \textit{Shannon entropy} \cite{mackay2003information} as a measure of uncertainty; i.e.\ high entropy (uncertain) signals given the GMM of the acoustic emission data \cite{AE} is illustrated in Figure~\ref{fig:entQ}.

In summary, as label information is limited by cost implications in practical SHM \cite{bull2019stsd}, active algorithms can be utilised to automatically administer the label budget, by selecting the most \textit{informative} data to be investigated -- such that the performance of predictive models is maximised \cite{bull2019machiningAL}.

\subsection{Dirichlet Process Mixture Models for Nonparametric clustering}
Dirichlet Process (DP) mixture models \cite{neal2000markov} offer another probabilistic framework to deal with limited labels as well as incomplete data \textit{a priori}. %
The DP is suggested as an (unsupervised) Bayesian algorithm for nonparametric clustering, used to perform inference online such that the need for extensive training-data (before implementing the SHM strategy) is mitigated \cite{rogers2019}. %
As such, unlike partially-supervised methods, labels are always an additional \textit{latent} variable (they are never observed); thus, the ground truth of $y_i$ is not known during inference. %
Label information has the potential to be incorporated, however; either within the SHM strategy \cite{rogers2019}, or at the algorithm level to define a semi-supervised DP \cite{vlachos2009unsupervised}. %

Conveniently, Bayesian properties of the DP allow the incorporation of prior knowledge and updates of belief, given the observed data. %
The aim is to avoid the need for comprehensive training-data, while retaining flexibility to include any available data formally as prior knowledge. %
Additionally, as there is a reduction in the number of user-tuned parameters, models can be implemented to perform powerful online learning with minimal \textit{a priori} input/knowledge, in terms of access to data or a physical model \cite{rogers2019}.

\subsubsection{Dirichlet Process Clustering}

A popular analogy to describe the DP (for clustering) considers a restaurant with an infinite number of tables \cite{aldous1985exchangeability} (i.e.\ clusters in $\mathscr{Y}$). Customers  -- resembling observations in $\mathscr{X}$ -- arrive and sit at one of the tables  (according to some probability) which are either occupied or vacant. As a table becomes more popular, the probability that customers join it increases. %
The seating arrangement can be viewed to represent a DP mixture. Importantly, the probability that a \textit{new} vacant table is chosen (over an existing table) is defined by a hyperparameter $\alpha$, associated with the DP. %
In consequence, $\alpha$ is sometimes referred to as the \textit{dispersion value} -- high values lead to an increased probability that new tables (clusters) are formed, while low values lead to less tables, as new tables are less likely to be initiated.

The analogy should highlight a useful property of DP mixtures: the number of clusters $K$ (i.e.\ tables) does not need to be defined in advance, instead, this is be determined by the model and the data (as well as $\alpha$) \cite{vlachos2009unsupervised}. %
As a result, the algorithm can be particularly useful when clustering SHM signals online, as the model can adapt and update, selecting the most appropriate value for $K$ as new information becomes available. %

\begin{figure}[pt]
    \centering
    \begin{subfigure}[b]{.49\textwidth}
      \centering
      \includegraphics[width=\linewidth]{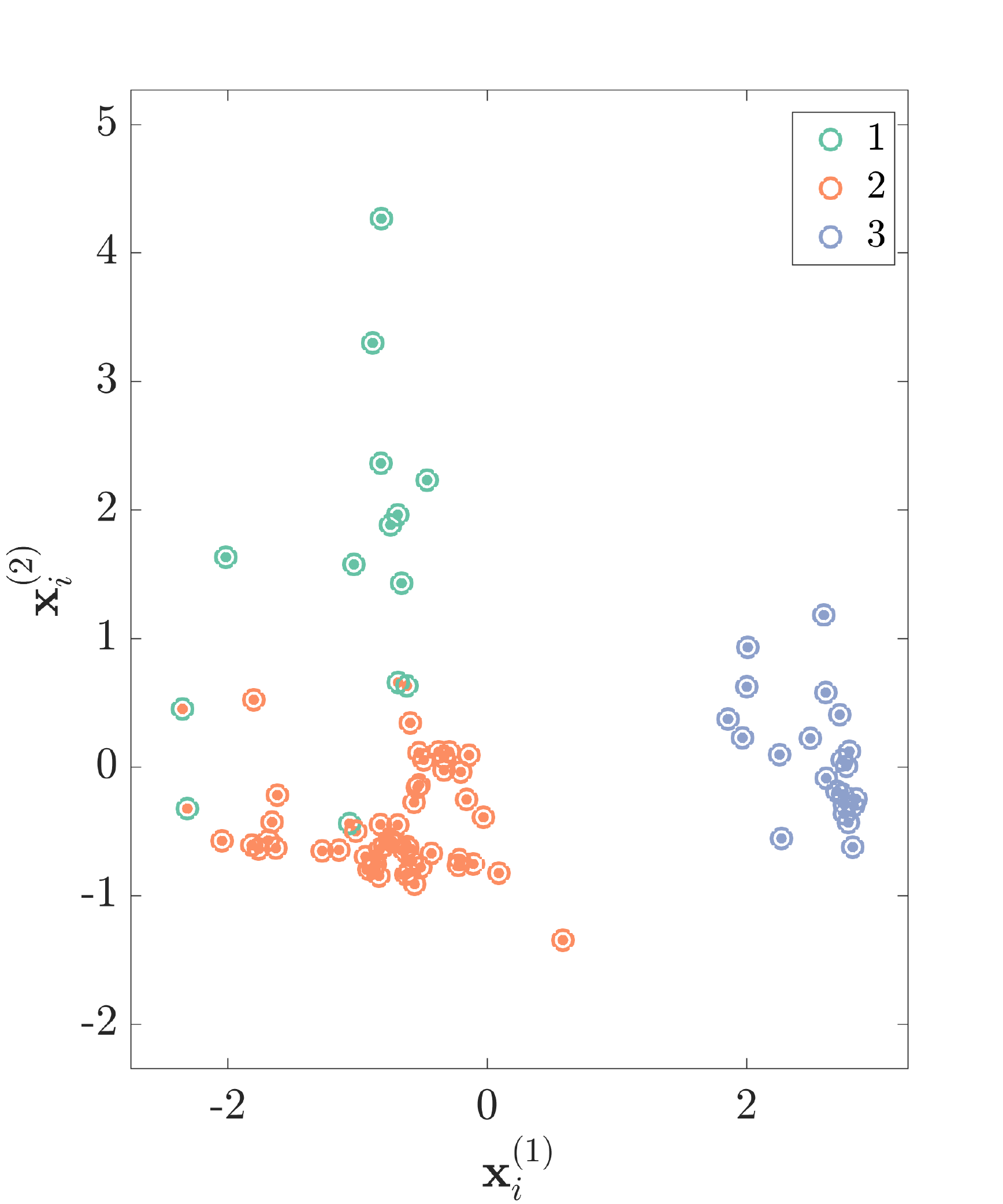}
      \caption{\label{a}}\label{fig:DPcl}
    \end{subfigure}
    \begin{subfigure}[b]{.49\textwidth}
      \centering
      \includegraphics[width=\linewidth]{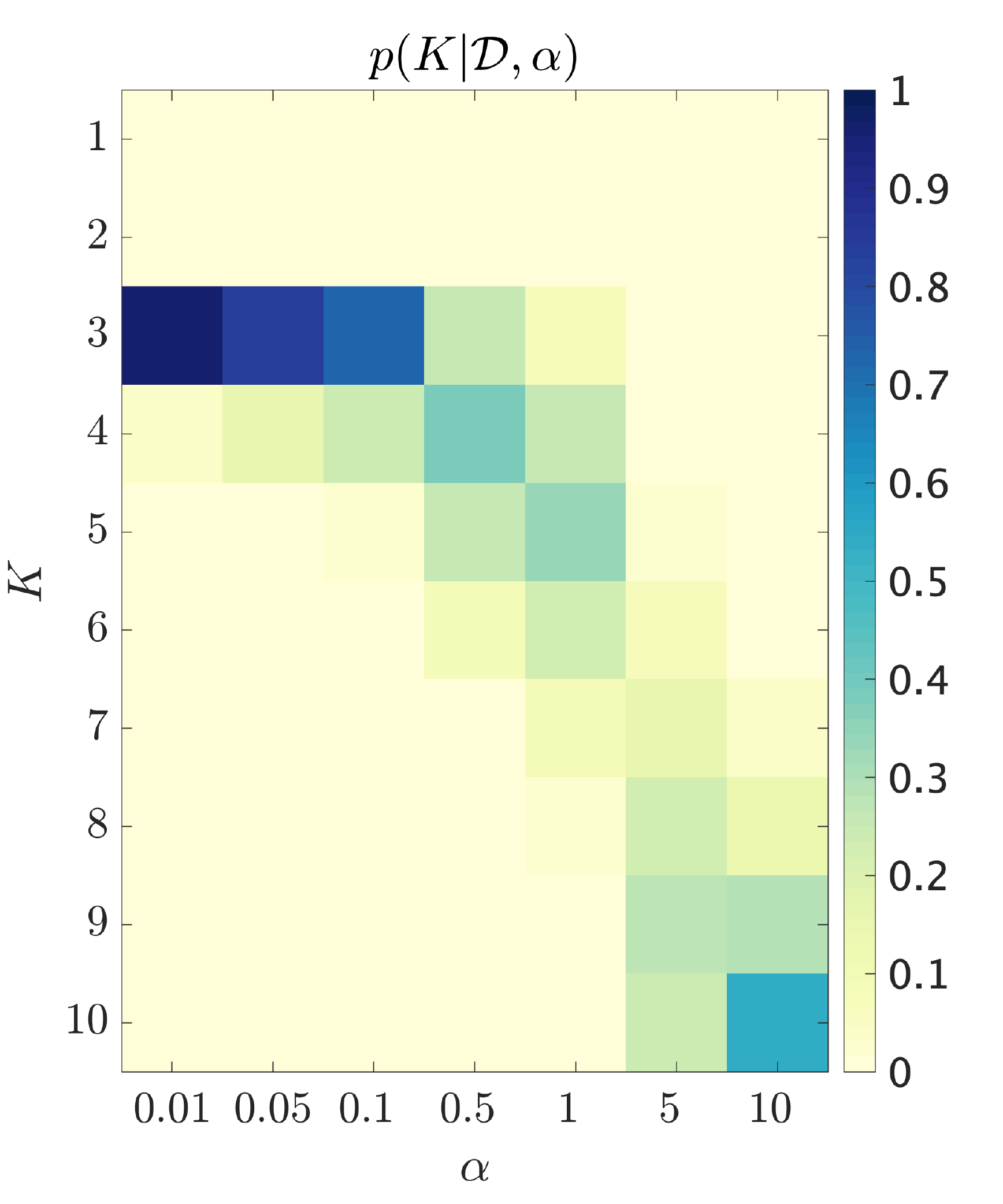}
      \caption{\label{b}}\label{fig:DPK}
    \end{subfigure}
    \caption{Unsupervised Dirichlet process Gaussian mixture model for the three-class AE data: (\subref{a}) unsupervised DP clustering, $\bullet/ \circ$ markers are the ground-truth/predicted values for $y_i$. (\subref{b}) predictive likelihood for the number of clusters $K$ given $\alpha$, i.e.\ $p(K|\mathcal{D},\alpha)$.}\label{fig:DP}
\end{figure}

To demonstrate, consider a mixture of Gaussian base-distributions; a conventional \textit{finite mixture} (a GMM) requires the number of components $K$ to be defined \textit{a priori}, as in the supervised Gaussian Mixture Model (GMM) with $K=3$, shown in Figures~\ref{fig:gmm_ss_eg} and~\ref{fig:al_ae}. %
As suggested by the analogy, a DP can be interpreted as an \textit{infinite} mixture, such that $K \rightarrow \infty $ \cite{rasmussen2000igmm}; this allows for the probabilistic inference of $K$ through the DP prior. An example DP-GMM for the same AE data \cite{AE} is shown in Figure~\ref{fig:DPcl}; the most likely number of components has been automatically found, $K = 3$, given the data and the model for $\alpha=0.1$. %
The effect of the \textit{dispersion} hyperparameter $\alpha$ can be visualised in Figure~\ref{fig:DPK}, which shows the posterior-predictive-likelihood of $K$ given the data for various values of $\alpha$. %
Considering that $K=3$, an appropriate hyperparameter range appears to be $0.01\leq\alpha\leq0.1$; although, as each class is clearly non-Gaussian, higher values of $K$ are arguably more appropriate to approximate the underlying density of the data. %
Interestingly, for low values of $\alpha$, three components appear significantly more likely to describe the data than two (or one). %

For SHM in practice, the implementation of the DP for online clustering means that an operator does not need to specify an expected number of normal, environmental or damage conditions (components $K$) in order to build the model, which can be difficult or impossible to define for a structure in operation \cite{rogers2019}.

\subsection{Transfer and Multi-task Learning}
Finally, methods for \textit{transfer} \cite{Gao2018,GARDNER2020106550,Jang2019} and \textit{multi-task} \cite{Ping2019,Yong2019} learning are proposed for inference with incomplete or limited training-data. In general terms, the idea for SHM applications is that valuable information might be transferred or shared, in some sense, between similar systems (via measured and/or simulated data). %
By considering \textit{shared} information, the performance of predictive models might improve, despite insufficient training observations \cite{Chakraborty2011,Ye2017,Dorafshan2018}. %
For example, consider wind turbines in an offshore wind-farm; one system may have comprehensively labelled measurements, investigated by the engineer, corresponding to a range of environmental effects; other turbines within the farm are likely to experience similar effects, however, the measured signals might be incomplete, with partial labelling or no labels at all. %

Various tools \cite{pan2009survey} offer frameworks to transfer \textit{different aspects} of shared information. %
For the methods discussed here, it is useful to define two objects \cite{GARDNER2020106550}:

\begin{itemize}
    \item A \textbf{Domain} $\mathscr{D} = \{\mathscr{X},p(\vec{x}_i)\}$ is an object that consists of a feature space $\mathscr{X}$ and a marginal probability distribution $p(\vec{x}_i)$ over a finite sample of feature data {$\left\{\vec{x}_i\right\}_{i=1}^{n} \in \mathscr{X}$}. \vspace{1em}
    \item A \textbf{Task} $\mathcal{T} = \{\mathscr{Y},f(\cdot)\}$ is a combination of a label space $\mathscr{Y}$ and a predictive model/ function $f(\cdot)$.
\end{itemize}

\begin{figure}[pt]
    \raggedright
    \begin{minipage}[b]{0.43\textwidth}
      \begin{subfigure}[b]{\linewidth}
        \centering
        \includegraphics[width=\linewidth]{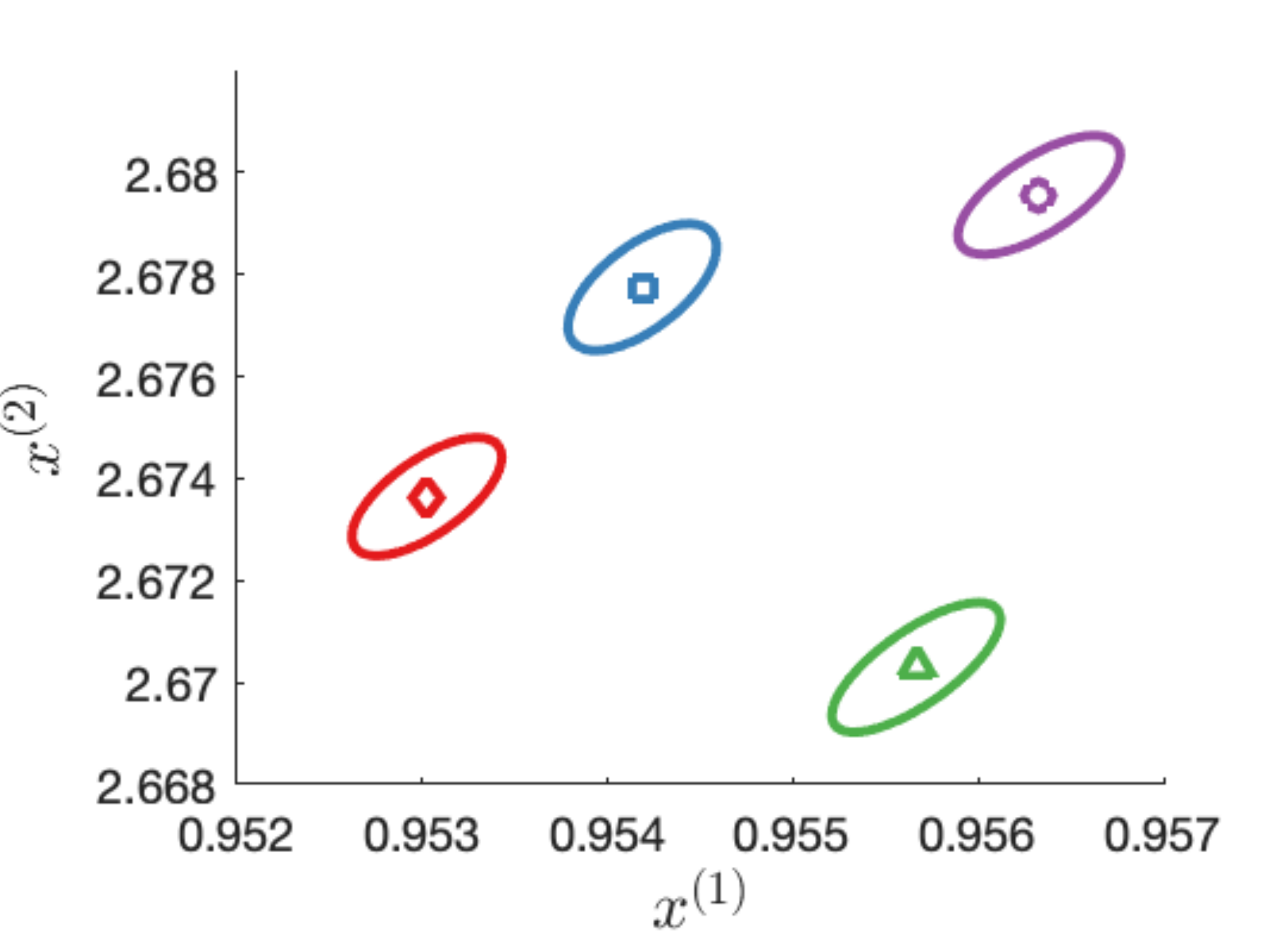}
        \caption{Source domain $\mathscr{D}_s$}\label{fig:Ds}
      \end{subfigure}\\[\baselineskip]
      \begin{subfigure}[b]{\linewidth}
        \centering
        \includegraphics[width=\linewidth]{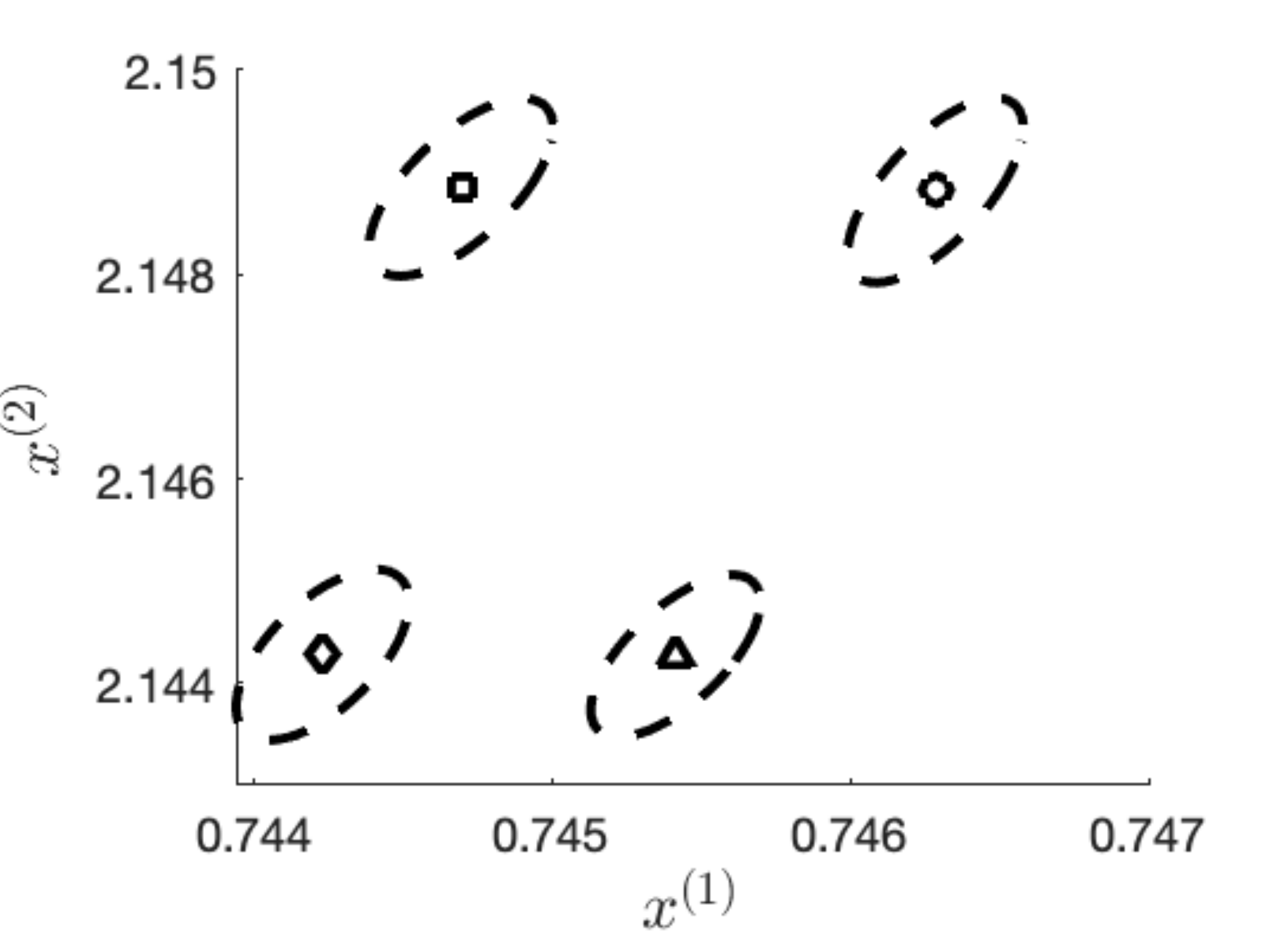}
        \caption{Target domain $\mathscr{D}_t$}\label{fig:Dt}
      \end{subfigure}
    \end{minipage}
    \hfill
    \begin{subfigure}[b]{0.54\textwidth}
        \centering
        \includegraphics[width=\linewidth]{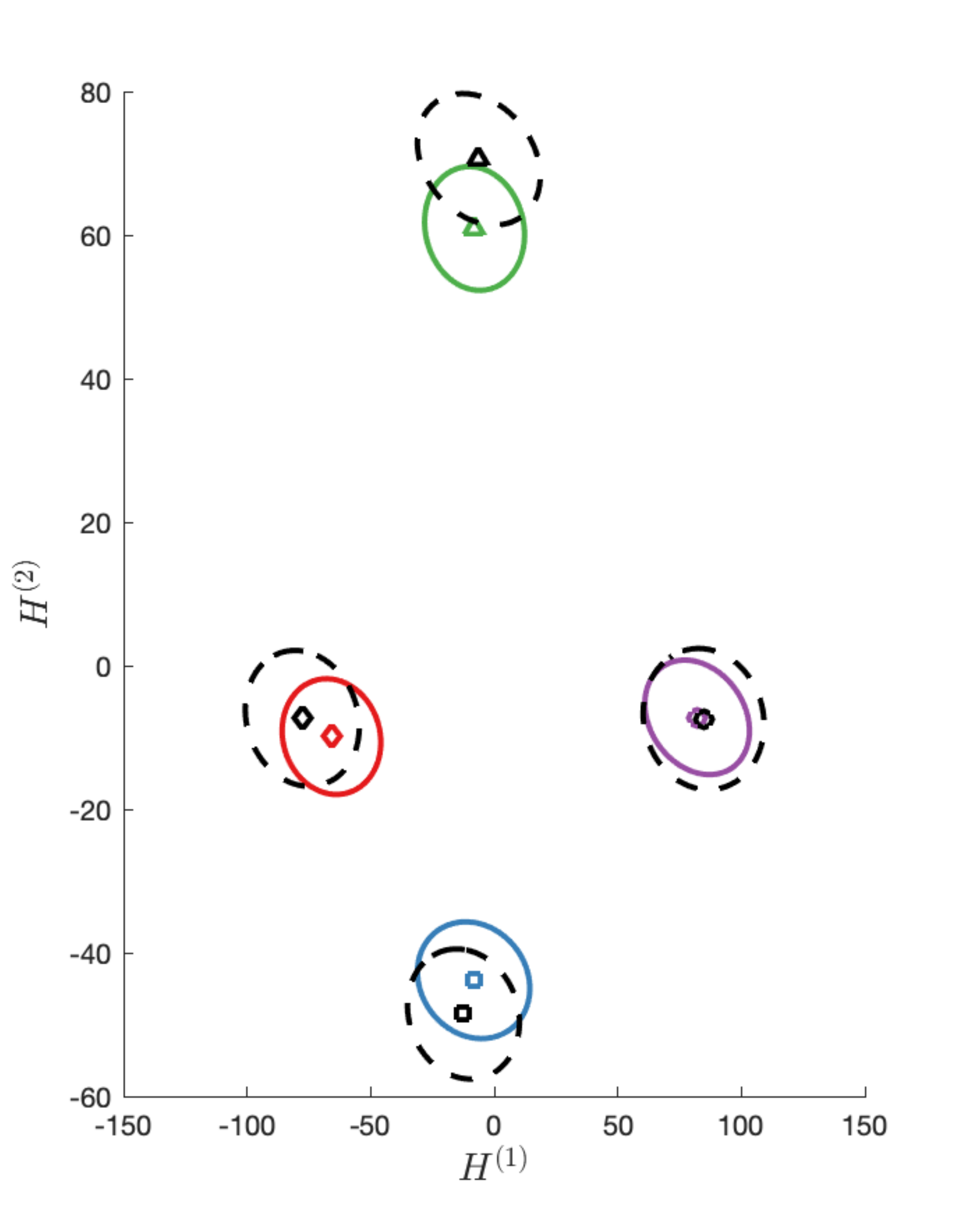}
      \caption{Latent space}\label{fig:Ls}
    \end{subfigure}    
    \caption{Visualisation of knowledge transfer via domain adaptation. Ellipses represent clusters of data -- coloured according to labels. (a) and (b) are the source and target domains respectively, in their original sample spaces. (c) shows the source and target data mapped into a shared, more consistent latent space.}\label{fig:Da}
  \end{figure}

\textit{Domain adaptation} is one approach to transfer learning, following a framework which maps the distributions from feature/label spaces (i.e.\ $\mathscr{X}$/$\mathscr{Y}$) associated with \textit{different} structures into a shared (more \textit{consistent}) space. The observations are typically \textit{labelled} for one structure only, therefore, a predictive model $f(\cdot)$ can be learnt, such that label information is \textit{transferred} between domains. %
The domain with labelled data is referred to as the \textit{source} domain $\mathscr{D}_s$ -- shown in Figure~\ref{fig:Ds} -- while the unlabelled data correspond to the \textit{target} domain $\mathscr{D}_t$ -- shown in Figure~\ref{fig:Dt}. %
Importantly, a classifier $f(\cdot)$ applied in the projected latent space of Figure~\ref{fig:Ls} should generalise to the target structure, despite missing label information. %

\textit{Multi-task learning} considers shared information from an alternative perspective. As with domain adaptation, knowledge from \textit{multiple domains} is used to improve tasks \cite{pan2009survey}; however, in this case, each domain is weighted equally \cite{zhang2018overview}. The goal is, therefore, to generate an improved predictive function $f(\cdot)$ across multiple tasks by utilising \textit{labelled} feature data from several different \textit{source domains}. %
This approach to inference is particularly useful when labelled training-data are insufficient across multiple tasks or systems. By considering the shared knowledge across various labelled domains, the amount of the training-data can, in effect, be increased. %

This work suggests \textit{kernelised Bayesian transfer learning} (KBTL) \cite{gonen2014kernelized} to model shared information. KBTL is a particular form of multi-task learning, which can be viewed as a method for \textit{heterogeneous} transfer; i.e.\ at least one feature space $\mathscr{X}_j$ for a domain $\mathscr{D}_j$ is not the same dimension as another feature space $\mathscr{X}_k$  (in the set of domains), such that $d_j \neq d_k$  \cite{GARDNER2020106550}. %
KBTL is a probabilistic method that performs two tasks: 1) finding a shared latent subspace for each domain and 2) inferring a discriminative classifier in the shared latent subspace in a Bayesian manner. It is assumed that there is a relationship between the feature space and the label space for each domain, and that all domains provide knowledge that will improve the predictive function $f(\cdot)$  for all domains \cite{GARDNER2020106550}. %

In practice, methods such as KBTL should be particularly useful for SHM, as the (labelled) training-data are often insufficient or incomplete across structures. If, through multi-task/transfer learning, tasks from \textit{different} structures can be considered together, this should increase the amount of information available to train algorithms. In turn, this should increase the performance of predictive models, utilising the \textit{shared} information between systems. %

\section{Directed Graphical Models}\label{s:DGM}
It will be useful to introduce basic concepts behind \emph{directed graphical models} (DGMs), as these will be used to (visually) introduce each probabilistic algorithm. The terminology here follows that of \cite{murphy}. %
Generally speaking, DGMs can be used to represent the joint distribution of the variables in a statistical model by making assumptions of \emph{conditional independence}. %
For these ideas to make sense, the \emph{chain rule} is needed; that is, the joint distribution of a probabilistic model can be represented as follows, using any ordering of the variables $\{X_1,X_2\ldots,X_V\}$: %

\begin{align}
    p(X_{1:V}) &= p(X_1)p(X_2\,|\,X_1)p(X_3\,|\,X_1,X_2)\ldots p(X_V\,|\,X_{1:V-1}) \label{eq:chain} \\ \nonumber
    X_{1:V} &\triangleq \{X_1,X_2\ldots,X_V\}
\end{align}

\noindent In practice, a problem with expression (\ref{eq:chain}) is that it becomes difficult to represent the conditional distribution $p(X_V\,|\,X_{1:V-1})$ as $V$ gets large. Therefore, to efficiently approximate large joint distributions, assumptions of conditional independence (\ref{eq:c_ind}) are critical. %
Specifically, conditional independence is denoted with $\bot$, and it implies that,

\begin{align}
    A\,\bot\,B\,|\,C \;&\longleftrightarrow\; p(A,B\,|\,C) =p(A\,|\,C)\,p(B\,|\,C)\label{eq:c_ind}
\end{align}

Considering these ideas, nodes in a graphical model can be used to represent variables, while edges represent conditional dependencies. %
For example, for the AE data (in Figures~\ref{fig:gmm_ss_eg}, \ref{fig:al_ae}, or \ref{fig:DPcl}), one can consider a random vector $\vec{x}_i$ to describe the (two-dimensional) measured features $\vec{x}_i = \left\{x^{(1)}_i, x^{(2)}_i\right\}$, and a random variable $y_i$ to represent the class label $\{1,2,3\}$. %
As a result, the joint distribution of an appropriate model might be $p\left(\vec{x}_i, y_i\right)$.
To simply matters, the features can be considered to be independent (an invalid but often acceptable assumption), i.e.\ $x^{(1)}_i\,\bot\,x^{(2)}_i\,|\,y_i$. This leads to the following approximation of distribution of the model (for a single observation):

\begin{align}
    p\left(\vec{x}_i, y_i\right) = p\left(x^{(1)}_i\;\lvert\;y_i\right)p\left(x^{(2)}_i\;\lvert\;y_i\right)p\left(y_i\right) \label{eq:DE}
\end{align}

\begin{figure}
    \centering
\begin{subfigure}[]{.4\textwidth}
    \centering
    \begin{tikzpicture}
    \tikzstyle{RV}=[circle, fill=white!100, minimum size = 3.5em, thick, draw = black!90, node distance = 4em]
    \tikzstyle{constant}=[circle, inner sep=0pt, fill=black!100, minimum size = 1.2mm, draw = black!80, node distance = 4em]
    \tikzstyle{plate}=[rectangle, rounded corners, draw=black!80, label={[yshift=17pt]south:#1}]{};
    \tikzstyle{connect}=[-latex, thick]
    
    \node[RV](Y)[]{$y_i$};
    \node[RV, fill=black!10](X1)[below left=of Y]{${x}^{(1)}_{i}$};
    \node[RV, fill=black!10](X2)[below right=of Y]{${x}^{(2)}_{i}$};
    
    \path (Y) edge [connect] (X1)
    (Y) edge [connect] (X2);
    \end{tikzpicture}
    \caption{\label{a}}\label{fig:dgm1}
\end{subfigure}
\begin{subfigure}[]{.4\textwidth}
    \centering
    \begin{tikzpicture}
    \tikzstyle{RV}=[circle, fill=white!100, minimum size = 3.5em, thick, draw = black!90, node distance = 4em]
    \tikzstyle{constant}=[circle, inner sep=0pt, fill=black!100, minimum size = 1.2mm, draw = black!80, node distance = 4em]
    \tikzstyle{plate}=[rectangle, thick, rounded corners, draw=black!50, label={[yshift=17pt, xshift=-4.5em]south east:#1}]{};
    \tikzstyle{connect}=[-latex, thick]
    
    \node[RV](Y)[]{$y_i$};
    \node[RV, fill=black!10](X1)[below=of Y]{${x}^{(j)}_{i}$};
     
    \node[plate=\small{$j = 1:d$}, inner sep=2em, fit= (X1)]{};
    \node[plate=\small{$i = 1:n$}, inner sep=4em, fit= (X1) (Y)]{};
    
    \path (Y) edge [connect] (X1);
    \end{tikzpicture}
    \caption{\label{b}}\label{fig:dgm2}
\end{subfigure}
\caption{Examples of directed graphical models (DGMs) based on the AE data. Shaded and unshaded nodes represent observed/latent variables respectively; arrows represent conditional dependencies; boxes represent plates.}\label{fig:DGM} 
\end{figure}
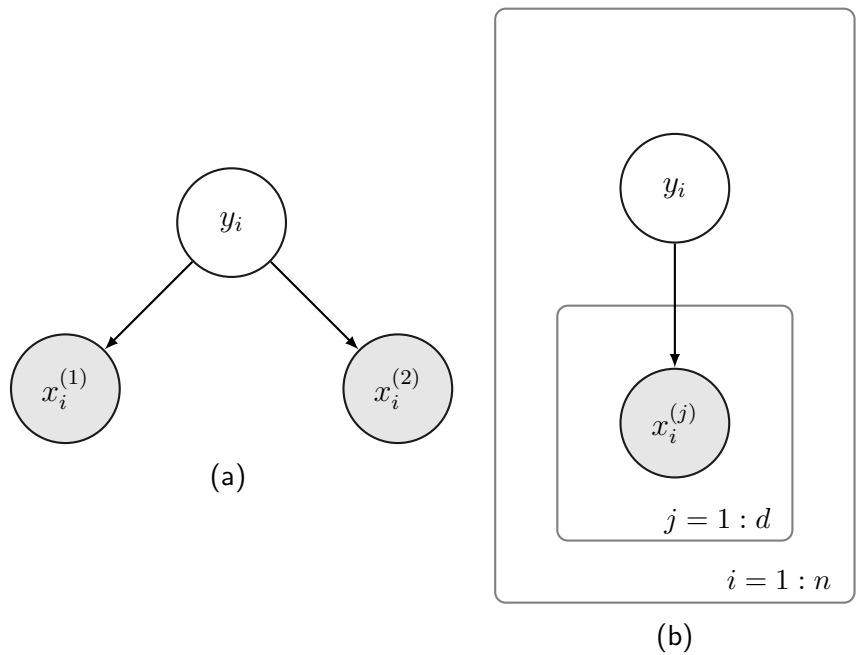

An appropriate distribution function $p(\cdot)$ can now be assigned to each of these densities (or masses). The DGM resulting from (\ref{eq:DE}) is plotted in in Figure~\ref{fig:dgm1}. In many cases, the features in $\vec{x}_i$ are the \textit{observed} variables (measured), while the labels $y_i$ are the \textit{latent} (or hidden) variables that one wishes to infer. To visualise this, the observed and latent variables are shown by shaded/unshaded nodes respectively in Figure~\ref{fig:dgm1}. %
For high-dimensional feature vectors (e.g.\ $d >> 2$), plates can be used to represent conditionally-independent variables and avoid a cluttered graph, as shown in Figure~\ref{fig:dgm2}. %
Another plate with ${i = \{1,\ldots,n\}}$ is included to represent \textit{independent and identically distributed} data, with $n$ observations. %
The DGM now represents the whole dataset, which is a matrix of observed variables $\vec{X} = \left\{\vec{x}_1,\ldots,\vec{x}_n\right\}$, and the vector of labels, denoted $\vec{y} = \left\{y_1,\ldots,y_n\right\}$. %
This assumptions implies that each sample was drawn independently from the same underlying distribution, such that the order in which data arrive makes no difference to the belief in the model, i.e.\ the likelihood of the dataset is, %

\begin{align}
  p\left(\vec{X}, \vec{y}\right) = \prod_{i=1}^{n}{p\left(x^{(1)}_i\;\lvert\;y_i\right)p\left(x^{(2)}_i\;\lvert\;y_i\right)p\left(y_i\right)} \label{eq:DE_N}
\end{align}

\noindent The corresponding DGM can be used to describe a (maximum likelihood) Na\"ive Bayes classifier -- a simplified version of the generative classifiers applied later in this work.

\section{Case Studies}\label{s:case_studies}
Semi-supervised, active, and multi-task learning, as well as DP clustering, are now demonstrated in case studies. A brief overview of the theory for each algorithm is provided, with the corresponding DGMs; for details behind each algorithm, the reader is referred to the SHM application papers \cite{onlineAL,BULL2020106653,rogers2019,GARDNER2020106550,imac_kbtl}. 

\subsection{Active learning with Gaussian Mixture Models}
A generative classifier is used to demonstrate probabilistic active learning. In this example -- originally shown in \cite{BULL2020106653} -- a Gaussian mixture model (GMM) is used to monitor streaming data from a motorway bridge, as if the signals were recorded online. The model defines a multi-class classifier, to aid both damage detection and identification, while limiting the number of (costly) system inspections.

\subsubsection{The directed graphical model}

As the data are being approximated by a Gaussian mixture model, when a new class $k$ is discovered from the streaming data (following inspection), it is assigned a Gaussian distribution -- Gaussian clusters like this can be visualised for the AE data in Figure~\ref{fig:gmm_ss_eg}. %
Note: the first DGM is explained in detail, to introduce the theory that is used throughout. %
The conditional distribution of the observations $\vec{x}_i$ given label $y_i = k$ is, therefore,

\begin{equation}\label{eq:c_likeli}
  p\left(\vec{x}_i \mid y_i = k\right) = \mathcal{N}\left(\vec{x}_i \,;\, \vec{\mu}_k, \vec{\Sigma}_k \right)
\end{equation}

(Semicolon notation $;$ is used to indicate that a function is parameterised by the variables that follow -- this is distinct from bar notation $\lvert$ which implies a conditional probability.) %
$k$ is used to index the class group, given the number of observed clusters at that time $k \in \left\{1,...,K\right\}$. As such, $\vec{\mu}_k$ is the mean (centre) and $\vec{\Sigma}_k$ is the covariance (scatter) of the cluster of data $\vec{x}_i$ with label $k$, for $K$ Gaussian base-distributions.

A discrete random variable is used to represent the labels $y_i$, which is categorically distributed, parameterised by a vector of \textit{mixing proportions} $\vec{\lambda}$, 

\begin{equation}\label{eq:c_prior}
  p\left(y_i \right) = \textrm{Cat}(y_i\,;\,\vec{\lambda})
\end{equation}

the mixing proportions can be viewed as a histogram over the label values, such that $\vec{\lambda} = \left\{\lambda_1,...,\lambda_K\right\}$ and $p(y_i=k) = P\left(y_i=k\right) = \lambda_k$.

The collected parameters of the model (from each component) are denoted by $\vec{\theta}$, such that ${\vec{\theta} = \left\{\vec{\Sigma},\vec{\mu},\vec{\lambda}\right\}} = \left\{\vec{\Sigma}_i,\vec{\mu}_i,\vec{\lambda}_i\right\}_{i=1}^K$; therefore, the joint distribution of the model could be written as,

\begin{align}
  p\left(\vec{x}_i,y_i\,;\,\vec{\theta}\right) = p\left(\vec{x}_i\,\lvert\,y_i\,;\,\vec{\theta}\right)p(y_i\,;\,\vec{\theta})
\end{align}

However, to consider a more \textit{complete} model, a Bayesian approach is adopted. That is, the parameters $\vec{\theta}$ themselves are considered to be random variables, and, therefore, they are included in the joint distribution (rather than simply parametersing it),

\begin{align}
  p\left(\vec{x}_i,y_i,\vec{\theta}\right) &= p\left(\vec{x}_i\,\lvert\,y_i,\vec{\theta}\right)p(y_i\,\lvert\,\vec{\theta})p\left(\vec{\theta}\right) \\
  &= p\left(\vec{x}_i\,\lvert\,y_i,\vec{\Sigma},\vec{\mu}\right)p\left(\vec{\Sigma},\vec{\mu}\right) p(y_i\,\lvert\,\vec{\lambda})p\left(\vec{\lambda}\right)
\end{align}

This perspective has various advantages; importantly, it allows for the incorporation of prior knowledge regarding the parameters via the \textit{prior distribution} $p\left(\vec{\theta}\right)$. Additionally, when implemented correctly, Bayesian methods lead to robust, self-regularising models \cite{rasmussen2001occam}.

To provide analytical solutions, it is convenient to assign conjugate (prior) distributions over the parameters $p\left(\vec{\theta}\right) = p\left(\vec{\Sigma},\vec{\mu})\,p(\vec{\lambda}\right)$. %
Here it is assumed that $\{\vec{\Sigma},\vec{\mu}\}$ are independent from $\vec{\lambda}$, to define two conjugate pairs; one associated with the observations $\vec{x}_i$ and another with the labels $y_i$. For the mean $\vec{\mu}_k$ and covariance $\vec{\Sigma}_k$, a conjugate (hierarchical) prior is the Normal Inverse Wishart (NIW) distribution,

\begin{equation}
  p(\vec{\mu}_k,\vec{\Sigma}_k) = \textmd{NIW}(\vec{\mu}_k,\vec{\Sigma}_k \,;\,\vec{m}_0, \kappa_0, \nu_0, \vec{S}_0) \label{eq:NIW}
\end{equation}

This introduces the \textit{hyperparameters} $\left\{\vec{m}_0, \kappa_0, \nu_0, \vec{S}_0\right\}$ associated with the prior, which can be interpreted as follows: $\vec{m}_0$ is the prior mean for the location of each class $\vec{\mu}_k$, and $\kappa_0$ determines the strength of the prior; $\vec{S}_0$ is (proportional to) the prior mean of the covariance, $\vec{\Sigma}_k$, and $\nu_0$ determines the strength of that prior \cite{murphy}. %
Considering that the streaming data will be normalised (online), it is reasonable that hyperparemeters are defined such that the prior belief states that each class is represented by a zero-mean and unit-variance Gaussian distribution. %

For the mixing proportions, the conjugate prior is a Dirichlet (Dir) distribution, parameterised by $\vec{\alpha}$, which encodes the prior belief of the mixing proportion (or weight) of each class. %
In this case, each class is assumed equally weighted \emph{a priori} for generality -- although, care should be taken when setting this prior, as it is application specific, particularly for streaming data \cite{onlineAL}.

\begin{align}
  p(\vec{\lambda}) &= \textmd{Dir}(\vec{\lambda}\,;\,\vec{\alpha}) \propto \prod^{K}_{k = 1}{\lambda_k}^{\alpha_k-1} \label{eq:dir}\\
  \vec{\alpha} &\triangleq \left\{\alpha_1,\ldots,\alpha_k\right\}
\end{align}

\begin{figure}
  \centering
  \begin{tikzpicture}
  \tikzstyle{RV}=[circle, fill=white!100, minimum size = 3.5em, thick, draw = black!90, node distance = 5em]
  \tikzstyle{constant}=[circle, inner sep=0pt, fill=black!100, minimum size = 1.2mm, draw = black!80, node distance = 4em]
  \tikzstyle{plate}=[rectangle, thick, rounded corners, draw=black!50, label={[yshift=17pt, xshift=-4.5em]south east:#1}]{};
  \tikzstyle{connect}=[-latex, thick]
    
  \node[RV, fill=black!10](X){$\vec{x}_{i}$};
  \node[RV](sigma)[left=of X]{$\vec{\Sigma}_{k}$};
  \node[RV](mu)[below=of sigma]{$\vec{\mu}_{k}$};
  \node[RV, fill=black!10](Y)[below=of X]{$y_i$};
  \node[RV](Pi)[right=of Y]{$\lambda_k$};
  \node[constant](alpha)[right=of Pi, label=below:$\vec{\alpha}$]{};
  
  \node[constant](sigma_0)[left=of sigma, label=left:$\vec{S}_0$]{};
  \node[constant](nu)[below = 1.6em of sigma_0, label=left:$\nu_0$]{};
  \node[constant](kappa_0)[left=of mu, label=left:$\kappa_0$]{};
  \node[constant](mu_0)[above = 1.6em of kappa_0, label=left:$\vec{m}_0$]{};
  
  \node[plate=\small{$i = 1:n$}, inner sep=2em, fit= (X) (Y)]{};
  \node[plate=\small{$k = 1:K$}, inner sep=2em, fit= (sigma) (mu)]{};
  \node[plate=\small{$k = 1:K$}, inner sep=2em, fit= (Pi)]{};
  \path (nu) edge [connect] (sigma)
  (sigma_0) edge [connect] (sigma)
  (kappa_0) edge [connect] (mu)
  (mu_0) edge [connect] (mu)
  (mu) edge [connect] (X)
  (sigma) edge [connect] (X)
  (Y) edge [connect] (X)
  (Pi) edge [connect] (Y)
  (alpha) edge [connect] (Pi)
  (sigma) edge [connect] (mu);
  \end{tikzpicture}
\caption{Directed graphical model for the GMM $p(\vec{x}_i, y_i, \vec{\theta})$ over the \textsl{labelled} data $\mathcal{D}_l$. As training data are supervised, both $\vec{x}_i$ and $y_i$ are observed variables. Shaded and white nodes are the observed and latent variables respectively; arrows represent conditional dependencies; dots represent constants (i.e.\ hyperparameters). Adapted from \protect\cite{bull_2019thesis}.}\label{fig:DG_GMM} 
\end{figure}
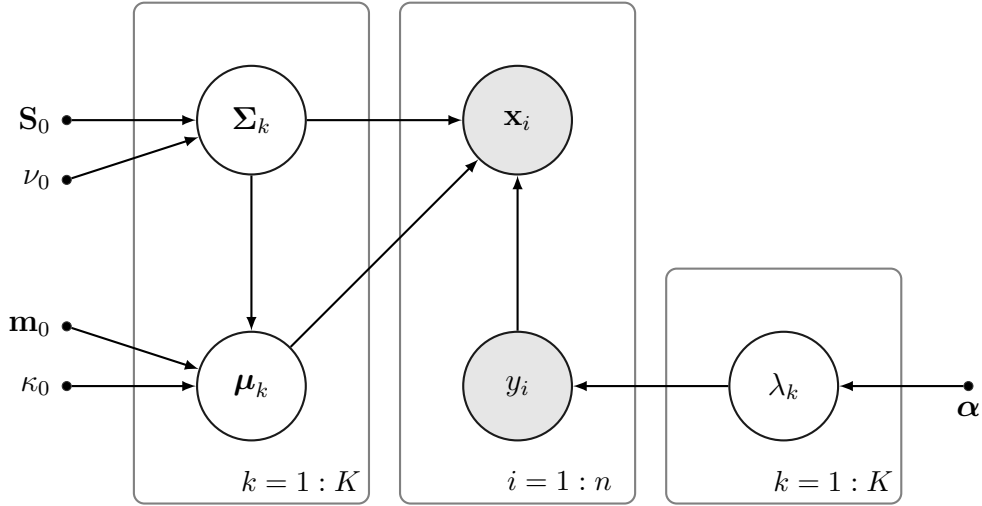

With this information, the joint distribution of the model $p(\vec{x}_i, y_i, \vec{\theta})$ can be approximated, such that $p(\vec{X}, \vec{y}, \vec{\theta}) = \prod_{i=1}^{n}{p(\vec{x}_i, y_i, \vec{\theta})}$. The associated DGM can be drawn, including conditional dependences and hyperparameters, for $n$ (supervised) training data in Figure~\ref{fig:DG_GMM}.

Having observed the labelled training data $\mathcal{D}_l = \left\{\vec{X},\vec{y}\right\}$ , the posterior distributions can be defined by applying Bayes' theorem to each conjugate pair -- where $\vec{X}_k$ denotes the observations $\vec{x}_i \in \vec{X}$ with the labels $y_i = k$,

\begin{align}
  p\left(\vec{\mu}_k,\vec{\Sigma}_k \mid \vec{X}_k, \right) &= \frac{p\left(\vec{X}_k \mid \vec{\mu}_k,\vec{\Sigma}_k\right) p\left(\vec{\mu}_k,\vec{\Sigma}_k\right)}{p(\vec{X}_k)} \label{eq:pos1}\\[1em]
  p\left(\vec{\lambda} \mid \vec{y} \right) &= \frac{p(\vec{y} \mid \vec{\lambda})p\left(\vec{\lambda}\right)}{p(\vec{y})}  \label{eq:pos2}
\end{align}

In general terms, while the prior $p(\vec{\theta})$ was the distribution over the parameters \textit{before} any data were observed, the posterior distribution $p(\vec{\theta}\mid\mathcal{D}_l)$ describes the parameters given the training data (i.e.\ conditioned on the training data). Conveniently, each of these have analytical solutions \cite{barber2012bayesian,murphy}.

\subsubsection{Active sampling}
To use the DGM to query informative data recorded from the motorway bridge, an initial model is learnt given a small sample of data recorded at the beginning of the monitoring regime. %
In this case, it should be safe to assume the labels $y_i=1$, which corresponds to the normal condition of the structure. %
As new (unlabelled) measurements arrive online, denoted $\tilde{\vec{x}}_i$, the model can be used to predict the labels \textit{under uncertainty}. %
The predictive equations are found by marginalising (integrating) out the parameters from the joint distribution (for each conjugate pair),

\begin{align}
  &p(\vec{\tilde{x}}_i \,|\, \tilde{y}_i = k, \mathcal{D}_l) = \int \int p(\vec{\tilde{x}}_i \,|\, \vec{\mu}_k,\vec{\Sigma}_k)\underbrace{p(\vec{\mu}_k,\vec{\Sigma}_k \,|\, \mathcal{D}_l)}_{\textrm{Eq.(\ref{eq:pos1})}}~d\vec{\mu}_k d{\vec{\Sigma}_k} \label{eq:pp1}\\[1ex]
  &p(\tilde{y}_i \,|\, \mathcal{D}_l) = \int p(\tilde{y}_i \,|\, \vec{\lambda}) \underbrace{p(\vec{\lambda} \,|\, \mathcal{D}_l)}_{\textrm{Eq.(\ref{eq:pos2})}}~d\vec{\lambda} \label{eq:pp2}
\end{align}

Again, due to conjugacy, these have analytical solutions \cite{murphy}. The posterior predictive equations (\ref{eq:pp1}) and (\ref{eq:pp2}) can be combined to define the posterior over the label estimates given unlabelled observations of the bridge,

\begin{equation}\label{bayes}
  p(\tilde{y}_i \,|\, \vec{\tilde{x}}_i,\mathcal{D}_l) = \frac{p(\vec{\tilde{x}}_i \,|\, \tilde{y}_i,\mathcal{D}_l)~p(\tilde{y}_i \,|\, \mathcal{D}_l)}{p(\vec{\tilde{x}}_i \,|\, \mathcal{D}_l)}
\end{equation}

Considering the predictive distribution (\ref{bayes}), labels that appear most uncertain can be investigated by the engineer. This observation is now labelled $\{\vec{x}_i, y_i\}$, thus extending the (supervised) training set $\mathcal{D}_l$. Two measures of uncertainty are considered: a) the marginal likelihood of the new observation given the model (the denominator of Equation (\ref{bayes})) and b) the entropy of the predicted label, given by,

\begin{equation}\label{entropy}
  H(\tilde{y}_i) = - \sum_{k=1}^{K}{p(\tilde{y}_i= k \,|\, \vec{\tilde{x}}_i,\mathcal{D}_l) \log{p(\tilde{y}_i= k \,|\, \vec{\tilde{x}}_i,\mathcal{D}_l)}}
\end{equation}

Queries with high entropy consider data at the boundary between two existing classes, while queries given low likelihood will select data that appear unlikely given the current model estimate. %
Visual examples of data that would be selected given these measures are shown in Figure \ref{fig:entQ} for high entropy, and Figure \ref{fig:likQ} for low likelihood.

\begin{figure}[pt]
  \centering
  \resizebox{\textwidth}{!}{%
  \begin{tikzpicture}[auto]
  \begin{footnotesize}
  \tikzstyle{decision} = [diamond, draw, text width=4em, text badly centered, inner sep=2pt]
  \tikzstyle{block} = [rectangle, draw, text width=10em, text centered, rounded corners, minimum height=4em]
  \tikzstyle{block2} = [rectangle, draw, text width=10em, text centered, rounded corners, minimum height=4em, fill=black!5]
  \tikzstyle{line} = [draw, -latex']
  \tikzstyle{cloud} = [draw=black!50, circle, node distance=3cm, minimum height=2em, text width=4em,text centered]
  \tikzstyle{point}=[draw, circle]
  
      \node [block2, node distance=4em] (start) {start:\\ initial training-set, $\mathcal{D}_l$};
      \node [point, below of=start, node distance=12mm] (point) {};
      \node [block, below of=point, node distance=12mm] (train) {train model\\ $p(\vec{\theta}\,|\, \mathcal{D}_l)$};
      \node [decision, below of=train, node distance=25mm] (new data) {new data?};
      \node [block2, left of=new data, node distance=40mm] (stop) {stop};
      \node [block, below of=new data, node distance=25mm] (update u) {update unlabelled set, $\mathcal{D}_u$};
      \node [cloud, left of=update u, node distance=40mm] (measured data) {measured data, batch size $B$};
      \node [block, right of=update u, node distance=42mm] (predict) {predict \\ $p(\tilde{y}_i \,|\, \vec{\tilde{x}}_i,\mathcal{D}_l), \forall \vec{\tilde{x}}_i \in \mathcal{D}_u$};
      \node [block, above of=predict, node distance=25mm] (query) {query $q_b$ informative data from $\mathcal{D}_u$};
      \node [cloud, right of=query, node distance=40mm] (annotate) {labels provided by the engineer};
      \node [block, above of=query, node distance=25mm] (update l) {update $\mathcal{D}_l$ to include new queried labels};
      
      \path [line] (start) -- (point);
      \path [line] (point) -- (train);
      \path [line] (train) -- (new data);
      \path [line] (new data) -- node {no}(stop);
      \path [line] (new data) -- node {yes}(update u);
      \path [line, dashed, draw=black!50] (measured data) -- (update u);
      \path [line] (update u) -- (predict);
      \path [line] (predict) -- (query);
      \path [line, dashed, draw=black!50] (annotate) -- (query);
      \path [line] (query) -- (update l);
      \path [line] (update l) |- (point);
  \end{footnotesize}
  \end{tikzpicture}
  }
  \caption{Flow chart to illustrate the online active learning process -- adapted from \protect\cite{onlineAL}.}
  \label{process}
\end{figure}
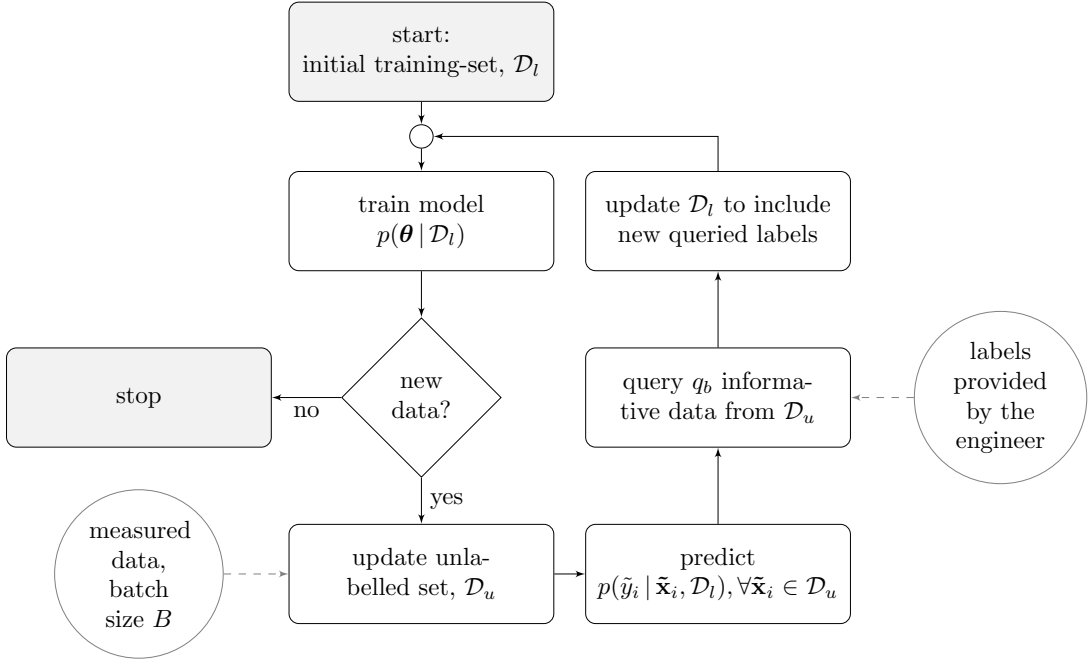

Figure \ref{process} demonstrates how streaming SHM signals might be queried using these uncertainty measures. %
The (unlabelled) data arrive online, in batches of size $B$; the data that appear most uncertain (given the current model) are investigated. %
The number of investigations per batch $q_b$ is determined by the label budget, which, in turn, is limited by cost implications. %
Once labelled by the engineer, these data can be added to $\mathcal{D}_l$ and used to update the classification model. %

\subsubsection{Z24 bridge dataset}
The Z24 bridge was a concrete highway bridge in Switzerland, connecting the villages of Koppigen and Utzenstorf. Before its demolition in 1998, the bridge was used for experimental SHM purposes \cite{SIMCES}. Over a twelve-month period, a series of sensors were used to capture dynamic response measurements, to extract the first four natural frequencies of the structure. Air/deck temperature, humidity and wind speed were also recorded \cite{OGz24}. There are a total of 3932 observations in the dataset. 

\begin{figure}[pt!]
\centering
\includegraphics[width=.8\linewidth]{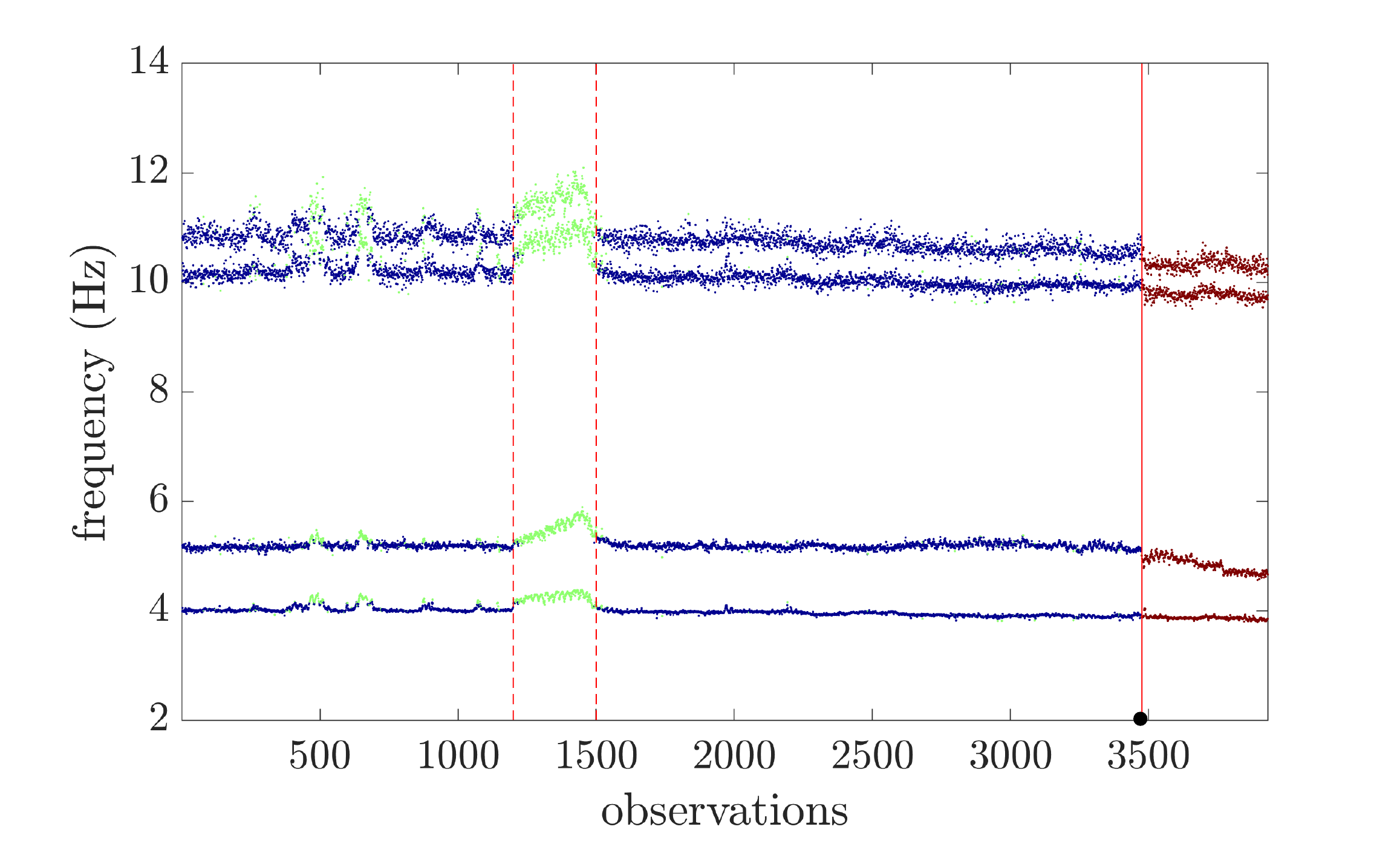}

\caption{Z24 bridge data, time history of natural frequencies, colours represent three classes of data: normal data (blue), outlying data due to environmental effects (green), and damage (red).}\label{z24data}
\end{figure}

Before demolition, different types of damage were artificially introduced, starting from observation 3476 \cite{robust_og}. The natural frequencies and deck temperature are shown in Figure~\ref{z24data}. Visible fluctuations in the natural frequencies can be observed in Figure~\ref{z24data}, for $1200 \leq n \leq 1500$, while there is little variation following the introduction of damage at observation 3476. %
It is believed that the asphalt layer in the deck experienced very low temperatures during this time, leading to increased structural stiffness. 

In the analysis, the four natural frequencies are the observation data, such that $\vec{x}_i \in \mathbb{R}^4$. The damage data are assumed to represent their own class, from observation 3476. Outlying observations within the remaining dataset are determined using the robust Minimum Covariance Determinant (MCD) algorithm \cite{fastmcd,robust_og}. %
In consequence, a three-class classification problem is defined, according to the colours in Figure~\ref{z24data}: normal data (blue), outlying data due to environmental effects (green), and damage (red), corresponding to $y_i \in \{1,2,3\}$ respectively.

Clearly, it is undesirable for an engineer to investigate the bridge following each data acquisition. Therefore, if active learning can provide an improved classification performance, compared to passive learning (random sampling) with the same sample budget, this demonstrates the relevance of active methods to SHM.

\subsubsection{Results: Active learning}
The model is applied \textit{online} to the frequency data from the Z24 bridge. %
To provide a online performance metric, the dataset is divided into two equal subsets: one is used for training and querying by the active learner $\{\mathcal{D}_l, \mathcal{D}_u\}$, the other is used as a distinct/independent test set. %
The $f_1$ score is used as the performance metric (throughout this work); this is a weighted average of precision and recall \cite{murphy}, with values between 0 and 1; a perfect score corresponds to $f_1 = 1$.
Precision (P) and recall (R) can be defined in terms of numbers of true positives ($TP$), false positives ($FP$) and false negatives ($FN$) for each class, $k \in Y$ \cite{murphy},

\begin{subequations}
        \begin{equation}
        P_k = \frac{TP_k}{TP_k + FP_k}
        \end{equation}
        \begin{equation}
        R_k = \frac{TP_k}{TP_k + FN_k}
        \end{equation}
\end{subequations}

The (macro) $f_1$ score is then defined by \cite{murphy},

\begin{subequations}\label{eq:f1}
        \begin{equation}
        f_{1,k} = \frac{2P_kR_k}{P_k + R_k}
        \end{equation}
        \begin{equation}
        f_{1} = \frac{1}{K} \sum_{k \in Y}{f_{1,k}}
        \end{equation}
\end{subequations}

\begin{figure}[pt]
  \centering
  \begin{subfigure}{.49\textwidth}
  \centering
  \includegraphics[width=\linewidth]{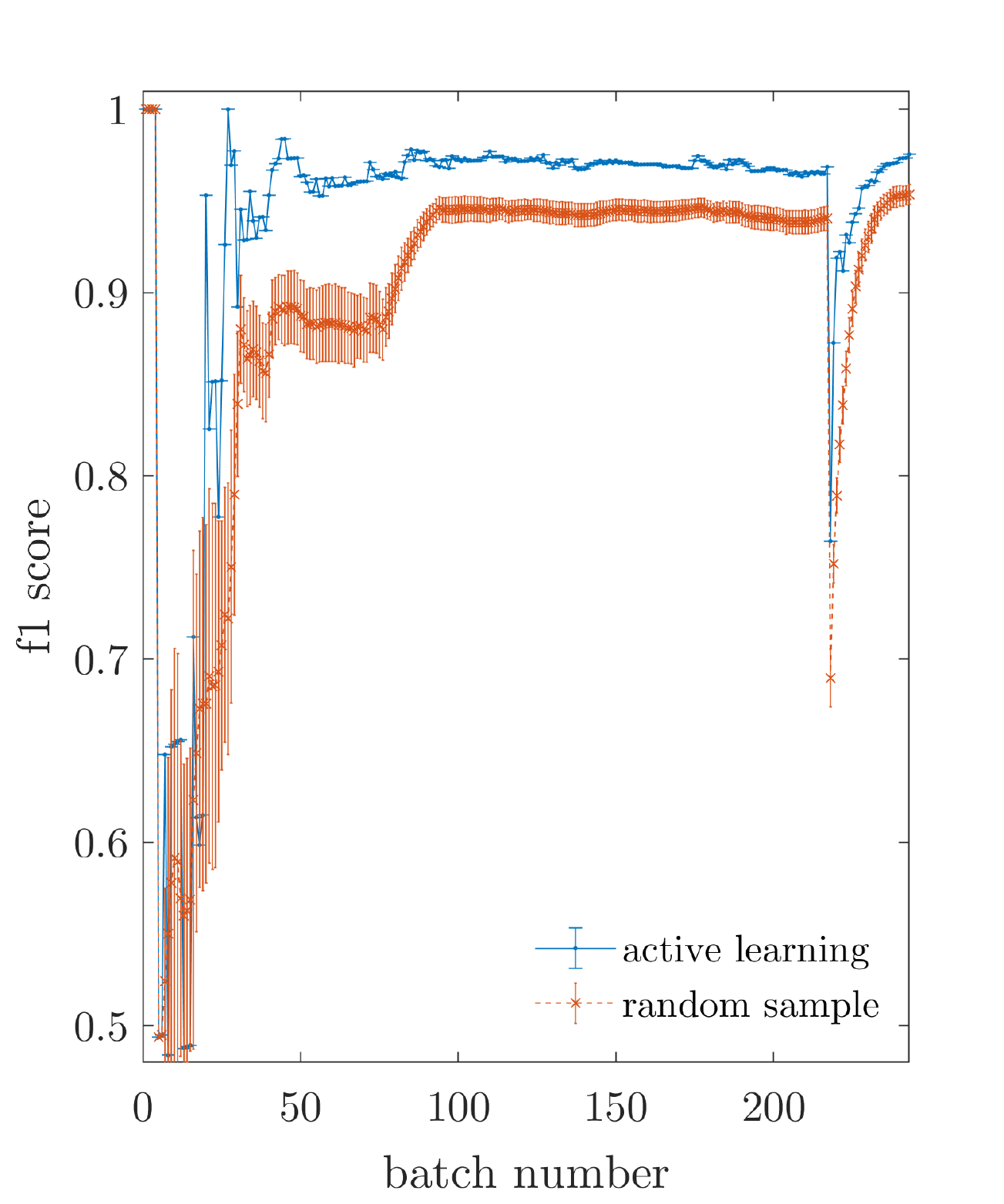}

  \caption{\label{a}}
  \label{z24ra}
  \end{subfigure}
  \begin{subfigure}{.49\textwidth}
  \centering
  \includegraphics[width=\linewidth]{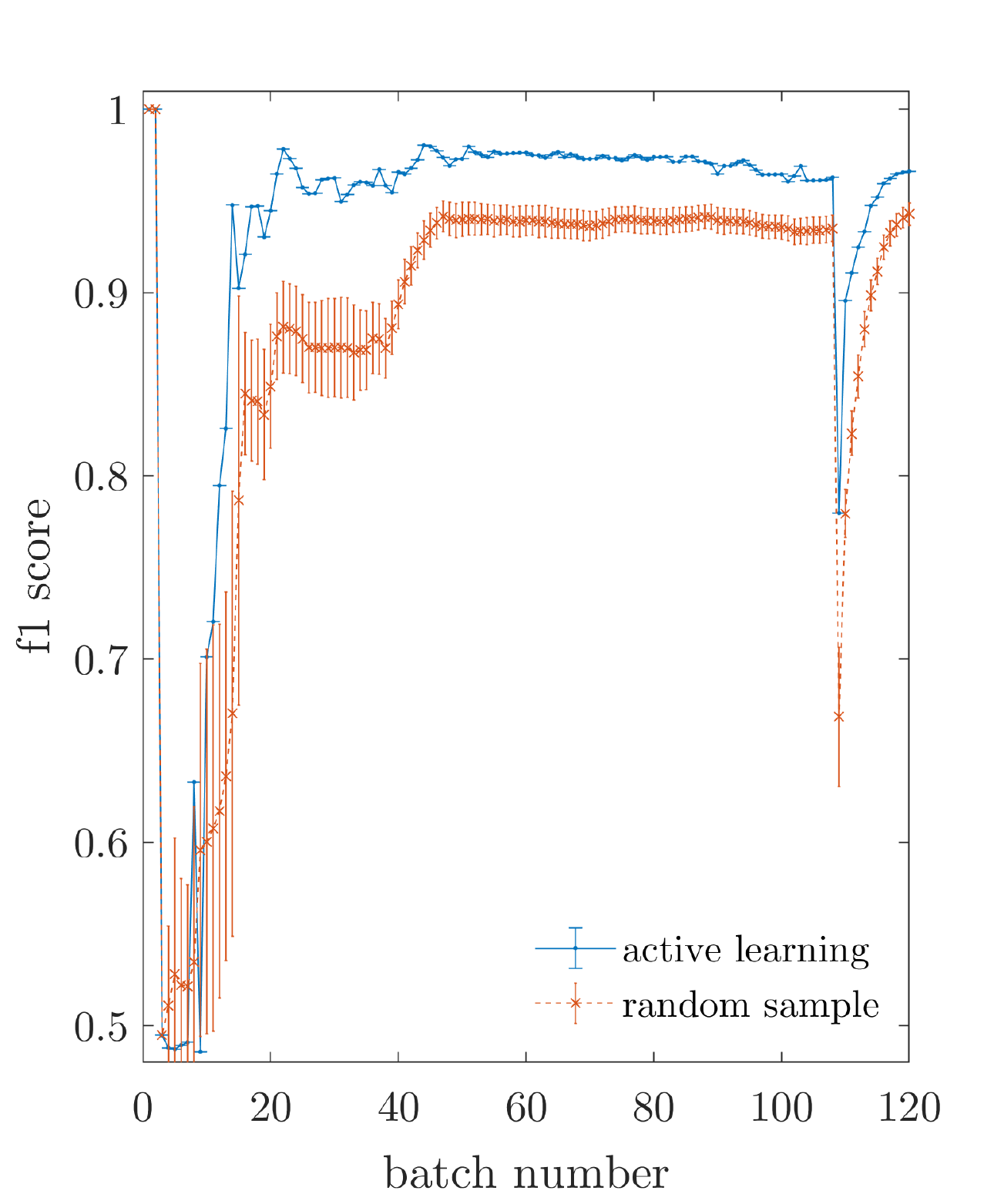}

  \caption{\label{b}}
  \label{z24rb}
  \end{subfigure}
  \caption{Online classification performance ($f_1$ score) for the Z24 data, for query budgets of (\subref{a}) 25\%; (\subref{b}) 12.5\% of the total dataset -- adapted from \protect\cite{onlineAL}.}\label{z24AL}
\end{figure}

Figure \ref{z24AL} illustrates improvements in classification performance when active learning is used to label 25\% and 12.4\% of the measured data. %
Active learning is compared to the \textit{passive} learning benchmark, where the same number of data are labelled according to a random sample, rather than uncertainty measures. %
Throughout the monitoring regime, if the GMM is used to select the training data, the predictive performance increases. %
Most notably, drops in the $f_1$ score (corresponding to new classes being discovered) are less significant when active learning is used to select data; particularly when class two (environmental effects) is introduced. %
This is because new classes are \emph{unlikely} given the current model, i.e.\ uncertainty measure (a). %
Intuitively, novel classes are discovered sooner via uncertainty sampling. %
For a range of query budgets and additional SHM applications refer to \cite{onlineAL}. %
Code and animations of uncertainty sampling for the Z24 data are available at \url{https://github.com/labull/probabilistic_active_learning_GMM}.

\subsection{Semi-supervised updates to Gaussian Mixture Models}
While active learning considered the unlabelled data $\mathcal{D}_u$ for querying, the observations only contribute to the model once labelled; in other words, once included in the labelled set $\mathcal{D}_l$. %
A semi-supervised model, however, can consider both the labelled \textit{and} unlabelled data when approximating the parameters. %
Therefore, ${\vec{\theta}}$ is estimated given \emph{both} labelled and unlabelled observations, such that the posterior becomes $p(\vec{\theta}\mid\mathcal{D}_l,\mathcal{D}_u)$. %
This is advantageous for SHM, \textit{unlabelled} observations can also contribute to the model estimate; reducing the dependance on costly supervised data. %

\begin{figure}
  \centering

  \begin{tikzpicture}
  \tikzstyle{RV}=[circle, fill=white!100, minimum size = 3.5em, thick, draw = black!90, node distance = 4em]
  \tikzstyle{constant}=[circle, inner sep=0pt, fill=black!100, minimum size = 1.2mm, draw = black!80, node distance = 2.5em]
  \tikzstyle{plate}=[rectangle, thick, rounded corners, draw=black!50, label={[yshift=17pt, xshift=-4.5em]south east:#1}]{};
  \tikzstyle{connect}=[-latex, thick]
  
  \node[RV, fill=black!10](X){$\vec{x}_{i}$};
  \node[RV](sigma)[left=of X]{$\vec{\Sigma}_{k}$};
  \node[RV](mu)[below=of sigma]{$\vec{\mu}_{k}$};
  \node[RV, fill=black!10](Y)[right=of X]{$y_i$};
  \node[RV, fill=black!10](Xu)[right=of mu]{$\vec{\tilde{x}}_{i}$};
  \node[RV](Yu)[right=of Xu]{$\tilde{y}_i$};
  \node[RV](Pi)[right =of Yu]{$\lambda_k$};

  \node[constant](alpha)[right=of Pi, label=below:$\vec{\alpha}$]{};
  \node[constant](sigma_0)[left=of sigma, label=left:$\vec{S}_0$]{};
  \node[constant](nu)[below = 1.6em of sigma_0, label=left:$\nu_0$]{};
  \node[constant](kappa_0)[left=of mu, label=left:$\kappa_0$]{};
  \node[constant](mu_0)[above = 1.6em of kappa_0, label=left:$\vec{m}_0$]{};
  
  \node[plate=\small{$i = 1:n$}, inner sep=1.6em, fit= (X) (Y)]{};
  \node[plate=\small{$i = 1:m$}, inner sep=1.6em, fit= (Xu) (Yu)]{};
  \node[plate=\small{$k = 1:K$}, inner sep=1.6em, fit= (sigma) (mu)]{};
  \node[plate=\small{$k = 1:K$}, inner sep=1.6em, fit= (Pi)]{};
  \path (nu) edge [connect] (sigma)
  (sigma_0) edge [connect] (sigma)
  (kappa_0) edge [connect] (mu)
  (mu_0) edge [connect] (mu)
  (mu) edge [connect] (X)
  (sigma) edge [connect] (X)
  (mu) edge [connect] (Xu)
  (sigma) edge [connect] (Xu)
  (Y) edge [connect] (X)
  (Yu) edge [connect] (Xu)
  (Pi) edge [connect] (Y)
  (Pi) edge [connect] (Yu)
  (alpha) edge [connect] (Pi)
  (sigma) edge [connect] (mu);
  \end{tikzpicture}
\caption{DGM of the semisupervised GMM, given the labelled $\mathcal{D}_l$ and unlabelled data $\mathcal{D}_u$. For the unsupervised set, $\vec{\tilde{x}}_i$ is the only observed variable, while $\tilde{y}_i$ is a latent variable. Adapted from \protect\cite{bull_2019thesis}.
}\label{fig:DGM_SS} 
\end{figure}
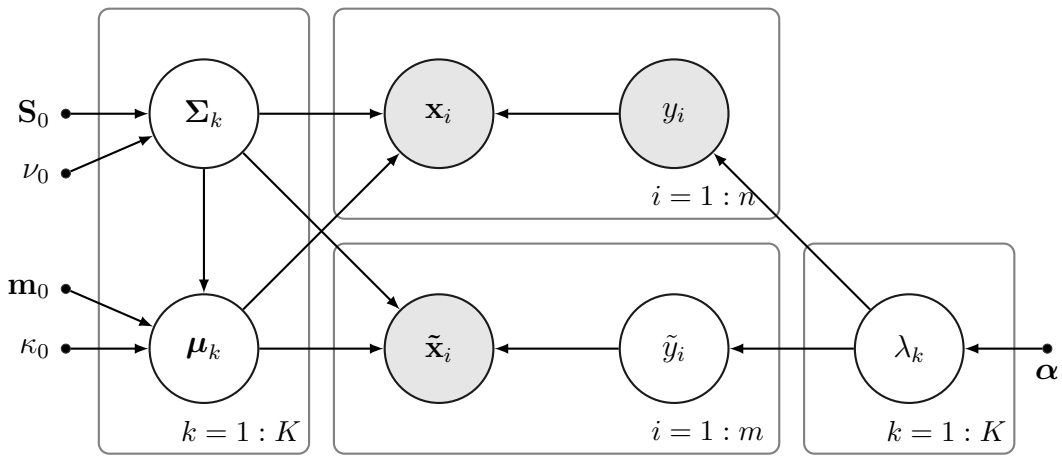

Continuing the probabilistic approach, the original DGM in Figure \ref{fig:DG_GMM} can be updated (relatively simply) to become semi-supervised -- shown in Figure~\ref{fig:DGM_SS}. The inclusion of $\mathcal{D}_u$ introduces another latent variable $\tilde{y}_i$, and, as a result, obtaining the posterior distribution over the parameters becomes less simple. %
One solution adopts an expectation maximisation (EM) approach \cite{dempster1977maximum}. %
The implementation here involves finding the maximum \textit{a posteriori} (MAP) estimate of the parameters $\vec{\hat{\theta}}$ (the mode of the full posterior distribution), while maximising the likelihood of the model. %
Specifically, from the joint distribution, and using Bayes' theorem, the MAP estimate of the parameters $\vec{\theta}$ given the labelled and unlabelled subsets is,

\begin{align}
  \nonumber \vec{\hat{\theta}}\;|\;\mathcal{D}\; &= \;\mathrm{argmax}_{\vec{\theta}}\left\{\frac{p(\mathcal{D} \,|\, \vec{\theta})p(\vec{\theta})}{p(\mathcal{D})} \right\}\\
  &= \;\mathrm{argmax}_{\vec{\theta}}\left\{\frac{p(\mathcal{D}_u \,|\, \vec{\theta})p(\mathcal{D}_l \,|\, \vec{\theta})p(\vec{\theta})}{p(\mathcal{D}_u,\mathcal{D}_l)} \right\}\label{eq:map}\\
  \mathcal{D} &\triangleq \mathcal{D}_u \cup \mathcal{D}_l \nonumber
\end{align}

Again, it is assumed that the data are i.i.d, so that $\mathcal{D}_l$ and $\mathcal{D}_u$ can be factorised. %
Thus, the marginal likelihood of the model (the denominator of equation~(\ref{eq:map})), considers both the labelled and unlabelled data -- this is referred to as the \textit{joint likelihood}, and it is the value that is maximised while inferring the parameters of the model through EM.

The EM algorithm iterates E and M steps until convergence in the joint (log) likelihood.  During each E-step, the parameters are fixed, and the unlabelled observations are classified using the current model estimate $p\left(\tilde{\vec{y}}\mid\tilde{\vec{X}}, \mathcal{D}\right)$. The M-step corresponds to finding the $\vec{\hat{\theta}}$, given the predicted labels from the E step \textit{and} the absolute labels for the supervised data. %
This involves some minor modifications to the conventional MAP estimates, such that the contribution of the unlabelled data is shared between classes, weighted according to the posterior distribution $p\left(\tilde{\vec{y}}\mid\tilde{\vec{X}}, \mathcal{D}\right)$ \cite{barber2012bayesian,BULL2020106653}. %
Pseudo-code is provided in Algorithm~\ref{EM}; Matlab code for the semi-supervised GMM is also available at \url{https://github.com/labull/semi_supervised_GMM}.

\begin{algorithm}[pt]
  \caption{\textsl{Semi-supervised EM for a Gaussian Mixture Model}}
    \label{EM}
    \SetAlgoLined
	\SetKwInOut{Input}{Input}\SetKwInOut{Output}{Output}
	\Input{~~Labelled data $\mathcal{D}_l$, unlabelled data $\mathcal{D}_u$}
	\Output{~~Semi-supervised MAP estimates of $\vec{\hat{\theta}} = \left\{\hat{\vec{\mu}},\hat{\vec{\Sigma}}\right\}$}
  \BlankLine
  \BlankLine
    \textit{Initilise} $\vec{\hat{\theta}}$ using the labelled data, $\vec{\hat{\theta}} = \textmd{argmax}_{\vec{\theta}}\left\{p(\vec{\theta}\,|\,\mathcal{D}_l)\right\}$\;
    \While{ the joint log-likelihood $\log\left\{p\left(\mathcal{D}_l,\mathcal{D}_u\right)\right\}$ improves}{
    \textit{E-step:} use the current model $\vec{\hat{\theta}}\mid \mathcal{D}$ to estimate class-membership for the unlabelled data $\mathcal{D}_u$, i.e.\ $p\left(\tilde{\vec{y}}\mid\tilde{\vec{X}}, \mathcal{D}\right)$\;
    \textit{M-step:} update the MAP estimate of  $\vec{\hat{\theta}}$ given the component membership for \textit{all} observations $\vec{\hat{\theta}} := \textmd{argmax}_{\vec{\theta}}\left\{p(\vec{\theta}\,|\,\mathcal{D}_l, \mathcal{D}_u)\right\}$\;}
\end{algorithm}

\subsubsection{Semi-supervised learning with the Gnat aircraft data}
A visual example of improvements to a GMM via semi-supervision was shown in Figure~\ref{fig:gmm_ss_eg}. To quantify potential advantages for SHM, the method is also applied to experimental data from aircraft experiments, originally presented in \cite{BULL2020106653}. %
For details behind the Gnat aircraft data, refer to \cite{partIII}. Briefly, during the tests, the aircraft was excited with an electrodynamic shaker and band-limited white noise. Transmissibilty data were recorded using a network of sensors distributed over the wing. Artificial damage was introduced by sequentially removing one of nine inspection panels in the wing. 198 measurements were recorded for the removal of each panel, such that the total number of (frequency domain) observations is 1782. %
Over the network of sensors, nine transmissibilties were recorded \cite{partIII}. Each transmissibility was converted to a one-dimensional novelty detector, with reference a distinct set of normal data, where all the panels were intact \cite{genetic}. %
Therefore, the data represent a nine-class classification problem, one class for the removal of each panel, such that $y_i = \{1,\ldots,9\}$. The measurements are nine-dimensional $\vec{x}_i \in \mathbb{R}^9$, each feature is a novelty index, representing one of nine transmissibilities. %

When applying semi-supervised learning, $1/3$ of the total data were set aside as an independent test-set. %
The remaining $2/3$ were used for training, i.e.\ $\mathcal{D} = \mathcal{D}_l \cup \mathcal{D}_u$. Of the training data $\mathcal{D}$, the number of labelled observations $n$ was increased (in 5\% increments) until all the observations are labelled. %
The results are compared to standard supervised learning for the same budget $n$. %

\begin{figure}[pt]
  \centering
  \includegraphics[width=\linewidth]{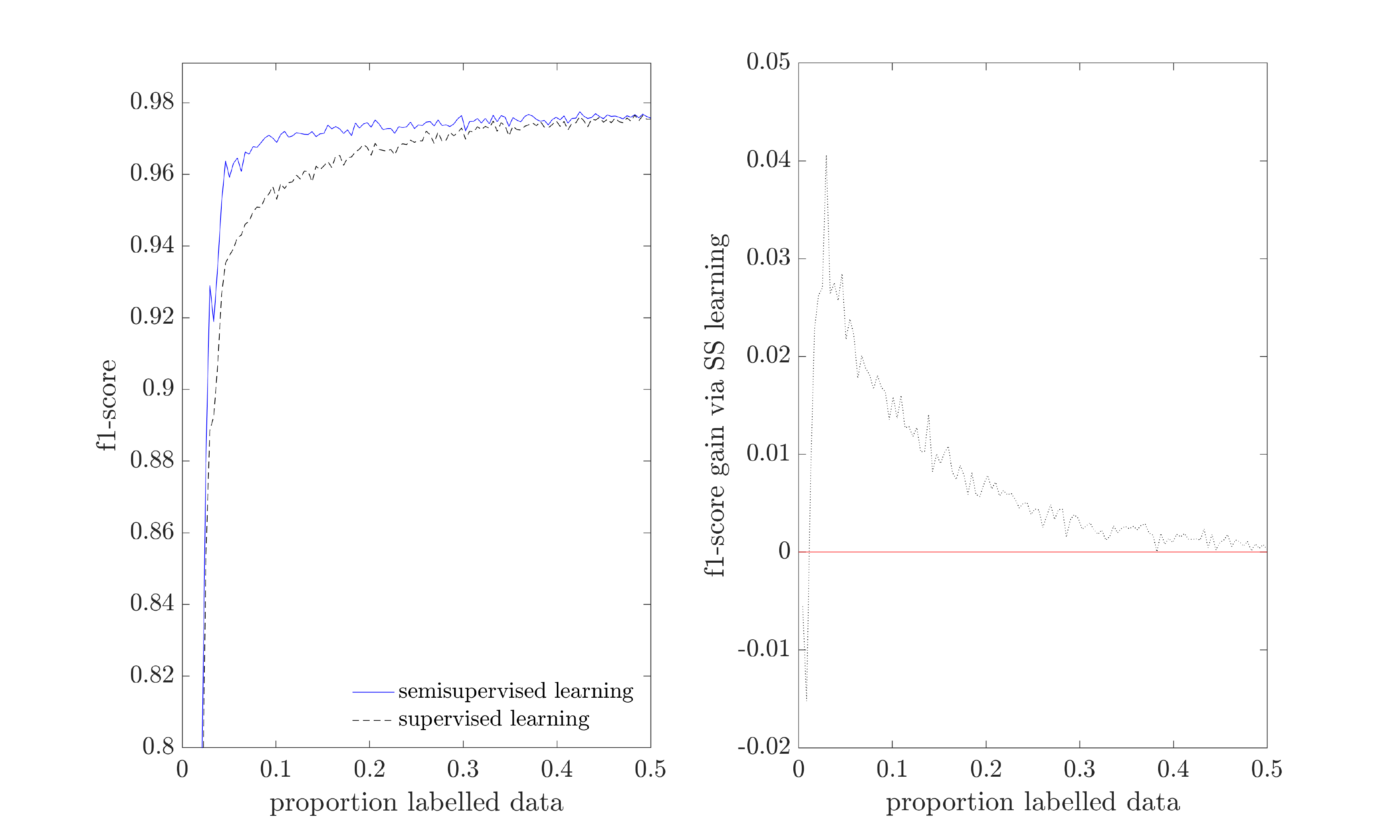}

  \caption{Classification performance ($f_1$ score) for the supervised GMM vs. the semi-supervised GMM. Left: $f_1$ for an increasing proportion of labelled data. Right: the gain in $f_1$ score through semi-supervised updates, the red line highlights zero-gain. Adapted from \protect\cite{BULL2020106653}.}\label{fig:res2a}
\end{figure}

The changes in the classification performance through semi-supervised updates are shown in Figure~\ref{fig:res2a}; inclusion of the unlabelled data consistently improves the $f_1$ score. %
For very low proportions of labelled data $<1.26\%$ ($m \gg n$), semi-supervised updates can decrease the predictive performance, this is likely due to the unlabelled data outweighing the labelled instances in the likelihood cost function. %
Notably, the maximum increase in the $f_1$ score is $0.0405$, corresponding to a 3.83\% reduction in the classification error for 2.94\% labelled data. %
Such improvements to the classification performance for for low proportions of labelled data should highlight significant advantages for SHM, reducing the dependence on large sets of costly supervised data.

\subsection{Dirichlet Process Clustering of Streaming Data}
Returning to the streaming data recorded from the Z24 bridge, an alternative perspective considers that labels are not needed to \textit{infer} the model. %
In this case, an \textit{unsupervised} algorithm could be used to cluster data online, and labels could be assigned to the resulting clusters \textit{outside} of the inference, within the wider SHM scheme -- as suggested by \cite{rogers2019}. %
However, if $y_i$ is unobserved for the purposes of inference, the number of class components $K$ becomes an additional latent variable, unlike the GMM from previous case studies. %

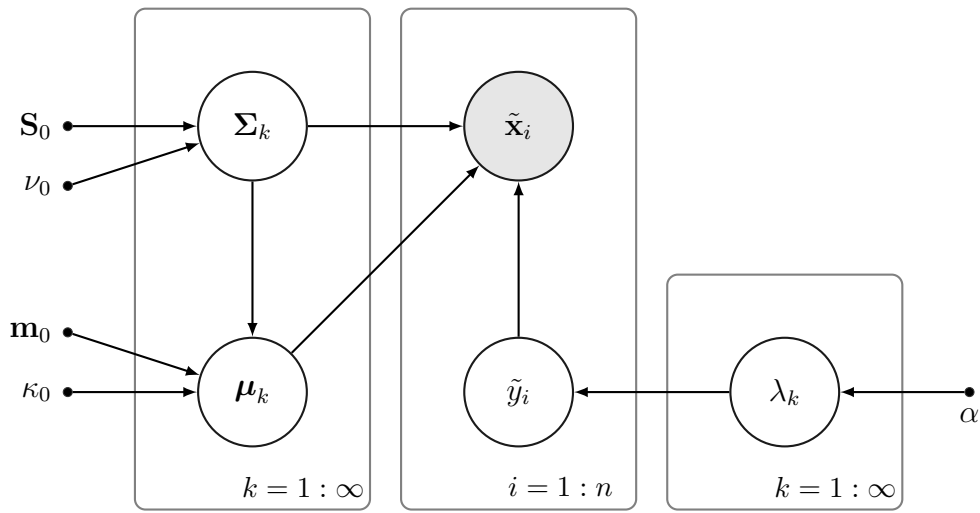
\begin{figure}[pt]
  \centering
  \begin{tikzpicture}
  \tikzstyle{RV}=[circle, fill=white!100, minimum size = 3.5em, thick, draw = black!90, node distance = 5em]
  \tikzstyle{constant}=[circle, inner sep=0pt, fill=black!100, minimum size = 1.2mm, draw = black!80, node distance = 4em]
  \tikzstyle{plate}=[rectangle, thick, rounded corners, draw=black!50, label={[yshift=17pt, xshift=-4.5em]south east:#1}]{};
  \tikzstyle{connect}=[-latex, thick]
    
  \node[RV, fill=black!10](X){$\tilde{\vec{x}}_{i}$};
  \node[RV](sigma)[left=of X]{$\vec{\Sigma}_{k}$};
  \node[RV](mu)[below=of sigma]{$\vec{\mu}_{k}$};
  \node[RV](Y)[below=of X]{$\tilde{y}_i$};
  \node[RV](Pi)[right=of Y]{$\lambda_k$};
  \node[constant](alpha)[right=of Pi, label=below:${\alpha}$]{};
  
  \node[constant](sigma_0)[left=of sigma, label=left:$\vec{S}_0$]{};
  \node[constant](nu)[below = 1.6em of sigma_0, label=left:$\nu_0$]{};
  \node[constant](kappa_0)[left=of mu, label=left:$\kappa_0$]{};
  \node[constant](mu_0)[above = 1.6em of kappa_0, label=left:$\vec{m}_0$]{};
  
  \node[plate=\small{$i = 1:n$}, inner sep=2em, fit= (X) (Y)]{};
  \node[plate=\small{$k = 1:\infty$}, inner sep=2em, fit= (sigma) (mu)]{};
  \node[plate=\small{$k = 1:\infty$}, inner sep=2em, fit= (Pi)]{};
  \path (nu) edge [connect] (sigma)
  (sigma_0) edge [connect] (sigma)
  (kappa_0) edge [connect] (mu)
  (mu_0) edge [connect] (mu)
  (mu) edge [connect] (X)
  (sigma) edge [connect] (X)
  (Y) edge [connect] (X)
  (Pi) edge [connect] (Y)
  (alpha) edge [connect] (Pi)
  (sigma) edge [connect] (mu);
  \end{tikzpicture}

\caption{DGM for the infinite Gaussian mixture model.}\label{fig:DGM_IGMM} 
\end{figure}

As aforementioned, the Dirichlet Process Gaussian Mixture Model (DPGMM) is one solution to this problem. %
The DPGMM allows for the probabilistic selection of $K$ through the a Dirichlet process prior. %
Initially, this involves defining a GMM in a Bayesian manner, using the same priors as before; however, by following \cite{rasmussen2000igmm}, it is possible to take the limit $K \rightarrow \infty$ to form an infinite Gaussian mixture model. %
Surprisingly, this concept can be shown through another simple modification to the first DGM in Figure~\ref{fig:DG_GMM}, leading to Figure~\ref{fig:DGM_IGMM}. %
The generative equations remain the same as (\ref{eq:c_likeli}), (\ref{eq:c_prior}), (\ref{eq:NIW}), and (\ref{eq:dir}).

A collapsed Gibbs sampler can be used to perform efficient online inference over this model \cite{neal2000markov}. %
Although potentially faster algorithms for variational inference exist \cite{blei2006variational}, it can be more practical to implement the Gibbs sampler when performing inference online. %
The nature of the Gibbs sampling solution is that each data point is assessed conditionally in the sampler, this allows the addition of new points online, rather than batch updates \cite{rogers2019}.

Within the Gibbs sampler, only components $k=\{1,\ldots,K+1\}$ need to be considered to cover the full set of possible clusters \cite{rasmussen2000igmm}. %
As with the GMM, there are two conjugate pairs in the model; therefore, the predictive equations remain analytical (leading to a \textit{collapsed} Gibbs sampler).
In brief/general terms: while fixing the parameters, the Gibbs scheme determines the likelihood of an observation $\tilde{\vec{x}}_i$ being sampled from an existing cluster $k = \{1,\ldots,K\}$, or an (as of yet) unobserved cluster $k = K+1$ (i.e.\ the prior). %
Given the posterior over the $K+1$ classes, the cluster assignment $\tilde{y}_i$ is sampled, and the model parameters are updated accordingly. %
This process is iterated until convergence. %

\subsubsection{Applications to the Z24 bridge data}
In terms of monitoring the streaming Z24 data, any new observations that relate to existing clusters will update the associated parameters. If a new cluster is formed, indicating novelty, this triggers an alarm. In this case, the cluster must contain at least 50 observations to indicate novelty; for details refer to \cite{rogers2019}. Upon investigating the structure, an appropriate description can be assigned to the unsupervised cluster index (outside of the inference). %
As before, the Z24 data are normalised in an online manner, thus, the hyperparemeters of the prior $p(\vec{\mu},\vec{\Sigma})$ encode this knowledge. %
The choice of the dispersion value $\alpha$, defining $p(\vec{\lambda})$, is more application dependent -- %
as discussed in the restaurant analogy, this determines the likelihood that new clusters will be generated. %
In \cite{rogers2019}, sensible values for online SHM applications were found to be between $0<\alpha<20$; for the Z24 data, this is set to $\alpha = 10$. %

\begin{figure}[pt]
    \centering
    \includegraphics[width=.8\textwidth]{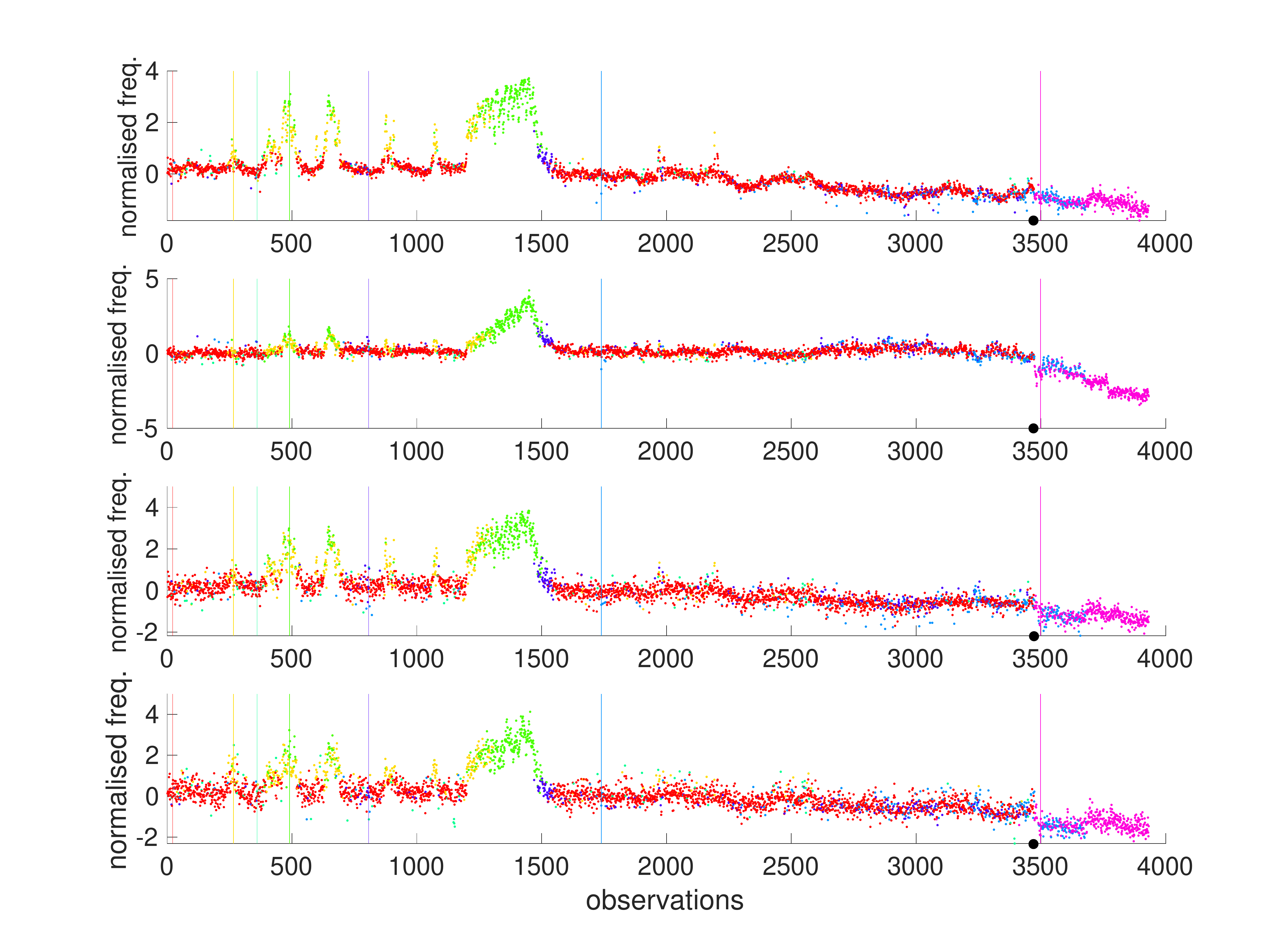}

    \caption{Figure showing online DP clustering applied to the Z24 bridge data using the first four natural frequencies as the features. Vertical lines indicate that a new cluster has been formed. Adapted from \protect\cite{rogers2019}.}
    \label{fig:time_cluster_z24}
\end{figure}

As with the active GMM, a small set of data from the start of the monitoring regime make up an initial training set. %
Figure~\ref{fig:time_cluster_z24} shows the algorithm progress for the streaming data. A normal condition cluster (red) is quickly established. As the temperature cools, three more cluster are created (orange, cyan and green) corresponding to the progression of freezing of the deck. Two additional clusters are also created: dark blue around point 800 and light blue close to point 1700. %
From inspection of the feature space  \cite{rogers2019}, it is hypothesised that the light blue cluster corresponds to a shift and rotation in the normal condition; therefore, this leads to another \emph{normal} cluster. As the corresponding normal data are now non-Gaussian, they are better approximated by two mixture components. %
Finally, the magenta cluster is created following two observations of damage, showing the ability of the DPGMM implementation to detect a change in behaviour corresponding to damage, as well as environmental effects. %

The DPGMM has automatically inferred seven clusters given the data and the model. While three classes were originally defined (as in the active and semi-supervised case), this representation is equally interpretable following system inspections to describe each component. %
Additionally, the DPGMM is likely to better approximate the underlying density, as each class of data can be described by a number of Gaussian components, rather than one. %
That is, in this case: three clusters describe the normal condition (blues and red), three clusters cover various environmental effects (orange, cyan and green), and one represents the damage condition (magenta). %

The results shown on the Z24 data demonstrate the ability of the online DP algorithm to deal with recurring environmental conditions while remaining sensitive to damage. %
The DPGMM is incorporated into an SHM system for online damage detection, and it is shown to categorise multiple damaged and undamaged states, while automatically inferring an appropriate number of mixture components $K$ in the mixture model. %
The method requires little user input, and it updates online with simple feedback to the user as to when inspection is likely required. If desired, the unsupervised clusters can be assigned meaningful descriptions, to be interpreted by the end user.

\subsection{Multi-task learning}
In the final case study, supervised data from different structures (each represented by their own domain) are considered simultaneously to improve the performance of an SHM task. %
In the following example, each domain $\mathscr{D}_t$ corresponds to supervised training data recorded from a different system; the task $\mathcal{T}$ corresponds to a predictive SHM model. %
By considering the data from a group (or population) of \textit{similar} structures in a latent space, the amount of training data can (in effect) be increased. %
Multi-task learning should be particularly useful in SHM, where training data are often incomplete for individual systems. %
If a predictive model can be improved by considering the data collected from various \textit{similar} structures, this should highlight the potential benefit of multi-task learning. %

\subsubsection{Kernelised Bayesian transfer learning}
Referring back to task $\mathcal{T}$ and domain $\mathscr{D}$ objects, it is assumed that there are $T$ (binary) classification tasks over the heterogeneous domains $\{\mathscr{D}_t\}_{t=1}^T$. %
In other words, the label space $\mathscr{Y}$ is consistent across all tasks (in this case, normal or damaged), while the feature space $\mathscr{X}_t$ can change dimensionality, potentially leading to $d_t \neq d_{t^{\prime}}$. %
For each task, there is an i.i.d.\ training set of observations $\vec{X}_t$ and labels $\vec{y}_t$, where $\vec{X}_t = \left\{ \vec{x}_i^{(t)} \in \mathbb{R}^{d_t} \right\}_{i=1}^{n_t}$ and $\vec{y}_t = \left\{ y^{(t)}_i \in \left\{ -1, +1 \right\} \right\}_{i=1}^{n_t}$. %
Each domain has a task specific kernel function $k_t$ to determine the similarities between observations and the associated kernel matrix $\vec{K}_t[i,j] = k_t\left(\vec{x}_i^{(t)}, \vec{x}_j^{(t)}\right)$, such that $\vec{K}_t \in \mathbb{R}^{n_t \times n_t}$. %
Note: when subscripts/superscripts are cluttered, square bracket notation is used to index matrices and vectors. %

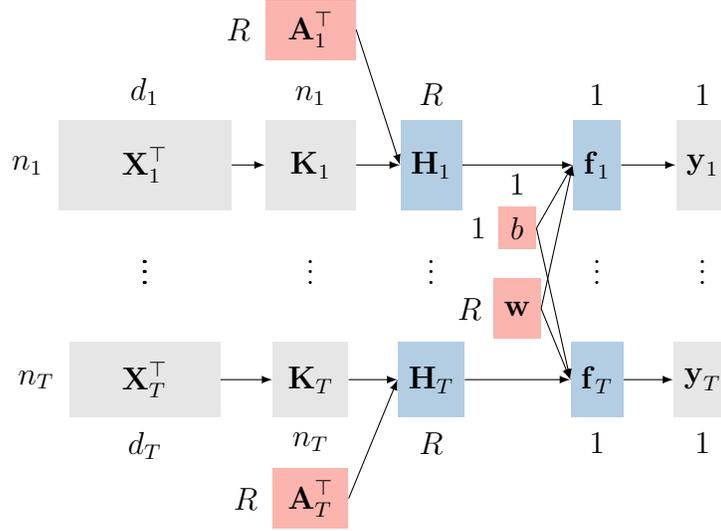
\begin{figure}[pt]
\centering
\begin{tikzpicture}
\node (n0) [] at (0cm,0cm) {$\vdots$};
\node (n1) [X1,above=.3cm of n0] {$\vec{X}_1^{\top}$};
\node (n3) [XT,below=1.5em of n0] {$\vec{X}_T^{\top}$};
\node (N1) [left=0.05cm of n1.west] {$n_1$};
\node (NT) [left=0.05cm of n3.west] {$n_T$};
\node (d1) [above=0.05cm of n1.north] {$d_1$};
\node (dT) [below=0.05cm of n3.south] {$d_T$};
\node (k0) [right=1.8cm of n0] {$\vdots$};
\node (k1) [K1,above=.3cm of k0] {$\vec{K}_1$};
\node (k2) [] {$\vdots$};
\node (k3) [KT,below=1.5em of k0] {$\vec{K}_T$};
\node (NN1) [above=0.05cm of k1.north] {$n_1$};
\node (NNT) [below=0.05cm of k3.south] {$n_T$};
\node (a1) [A1,above=2.3cm of k0] {$\vec{A}_1^{\top}$};
\node (a3) [AT,below=2.3cm of k0] {$\vec{A}_T^{\top}$};
\node (r1) [left=0.05cm of a1.west] {$R$};
\node (rT) [left=0.05cm of a3.west] {$R$};
\node (h0) [right=1.2cm of k0] {$\vdots$};
\node (h1) [H1,above=.3cm of h0] {$\vec{H}_1$};
\node (h2) [] {$\vdots$};
\node (h3) [HT,below=1.5em of h0] {$\vec{H}_T$};
\node (rr1) [above=0.05cm of h1.north] {$R$};
\node (rrT) [below=0.05cm of h3.south] {$R$};
\node (bw0) [right=2.4cm of k0] {};
\node (bw1) [B,above=.1em of bw0] {$b$};
\node (bw2) [W,below=.1em of bw0.south] {$\vec{w}$};
\node (bd) [above=0.01cm of bw1.north] {$1$};
\node (bn) [left=0.01cm of bw1.west] {$1$};
\node (wn) [left=0.01cm of bw2.west] {$R$};
\node (f0) [right=1.8cm of h0] {$\vdots$};
\node (f1) [f1,above=.3cm of f0] {$\vec{f}_1$};
\node (f2) [] {$\vdots$};
\node (f3) [fT,below=1.5em of f0] {$\vec{f}_T$};
\node (dd1) [above=0.05cm of f1.north] {$1$};
\node (ddT) [below=0.05cm of f3.south] {$1$};
\node (y0) [right=1cm of f0] {$\vdots$};
\node (y1) [y1,above=.3cm of y0] {$\vec{y}_1$};
\node (y2) [] {$\vdots$};
\node (y3) [yT,below=1.5em of y0] {$\vec{y}_T$};
\node (dy1) [above=0.05cm of y1.north] {$1$};
\node (dyT) [below=0.05cm of y3.south] {$1$};
\draw[->,>=latex] (n1.east) -- (k1.west);
\draw[->,>=latex]  (n3.east) -- (k3.west);
\draw[->,>=latex]  (k1.east) -- (h1.west);
\draw[->,>=latex]  (k3.east) -- (h3.west);
\draw[->,>=latex]  (a1.east) -- (h1.west);
\draw[->,>=latex]  (a3.east) -- (h3.west);
\draw[->,>=latex]  (h1.east) -- (f1.west);
\draw[->,>=latex]  (h3.east) -- (f3.west);
\draw[->,>=latex]  (f1.east) -- (y1.west);
\draw[->,>=latex]  (f3.east) -- (y3.west);
\draw[->,>=latex]  (bw1.east) -- (f1.west);
\draw[->,>=latex]  (bw1.east) -- (f3.west);
\draw[->,>=latex]  (bw2.east) -- (f1.west);
\draw[->,>=latex]  (bw2.east) -- (f3.west);
\end{tikzpicture}
\caption{Visualisation of KBTL -- adapted from \protect\cite{gonen2014kernelized}.}\label{fig:kbtl_flow}
\end{figure}

Figure~\ref{fig:kbtl_flow} is useful to visualise KBTL. The model can be split into two main parts: (i) the first projects data from different tasks into a shared subspace using kernel-based dimensionality reduction, (ii) the second performs \textit{coupled} binary classification in the shared subspace, using common classification parameters. In terms of notation, the kernel embedding for each domain $\vec{K}_t$ is projected into a shared latent subspace by an optimal projection matrix $\vec{A}_t \in \mathbb{R}^{n_t \times R}$, where $R$ is the dimensionality of the subspace. %
Following projection, there is a representation of each domain in the shared latent subspace, $\left\{ \vec{H}_t = \vec{A}_t^{\top}\vec{K}_t\right \}_{t=1}^{T}$. %
In this shared space, a \emph{coupled} discriminative classifier is inferred for the projected data from each domain $\left\{ \vec{f}_t = \vec{H}_t^{\top}\vec{w} + \vec{1}b\right \}_{t=1}^{T}$. This implies the same set of parameters $\left\{ \vec{w}, b\right\}$ are used across all tasks. %

In a Bayesian manner, prior distributions are associated with the parameters of the model. For the $n_t \times R$ task-specific projection matrices, $\vec{A}_t$, there is an $n_t \times R$ matrix of priors, denoted $\vec{\Lambda}_t$. For the weights of the coupled classifier, the prior is $\vec{\eta}$, and for the bias $b$ the prior is $\gamma$. These are standard priors given the parameter types in the model -- for details refer to \cite{gonen2014kernelized}. Collectively, the priors are $\vec{\Xi} = \left\{ \left\{\vec{\lambda}_t\right\}_{t=1}^{T}, \vec{\eta}, \gamma \right\}$ and the latent variables are $\vec{\Theta} =\left\{\left\{\vec{H}_t,\vec{A}_t, \vec{f}_t \right\}_{t=1}^{T}, \vec{w}, b \right\}$; the observed variables (training data) are given by $\left\{\vec{K}_t, \vec{y}_t \right\}_{t=1}^{T}$. %

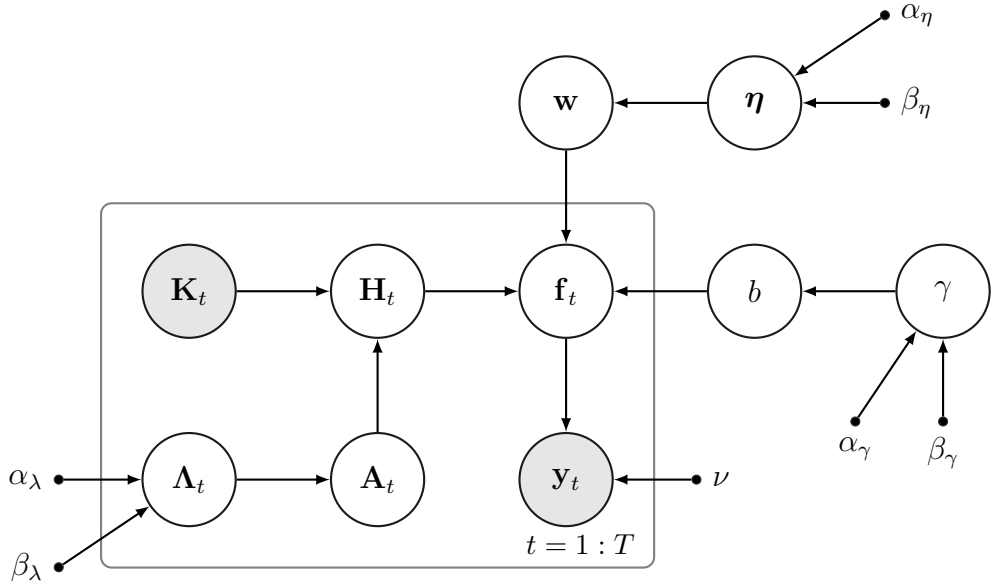
\begin{figure}[pt]
  \centering
  \begin{tikzpicture}
  \tikzstyle{RV}=[circle, fill=white!100, minimum size = 3em, thick, draw = black!90, node distance = 3em]
  \tikzstyle{constant}=[circle, inner sep=0pt, fill=black!100, minimum size = 1.2mm, draw = black!80, node distance = 2.5em]
  \tikzstyle{plate}=[rectangle, thick, rounded corners, draw=black!50, label={[yshift=17pt, xshift=-4.5em]south east:#1}]{};
  \tikzstyle{connect}=[-latex, thick]
    
  \node[RV](lambda){$\vec{\Lambda}_t$};
  \node[RV](A)[right=of lambda]{$\vec{A}_t$};
  \node[RV](H)[above=of A]{$\vec{H}_t$};
  \node[RV, fill=black!10](K)[above=of lambda]{$\vec{K}_t$};
  \node[RV, fill=black!10](y)[right=of A]{$\vec{y}_t$};
  \node[RV](f)[right=of H]{$\vec{f}_t$};
  \node[RV](w)[above=of f]{$\vec{w}$};
  \node[RV](eta)[right=of w]{$\vec{\eta}$};
  \node[RV](b)[right=of f]{$b$};
  \node[RV](gamma)[right=of b]{$\gamma$};
  
  \node[constant](alpha_l)[left=of lambda, label=left:$\alpha_\lambda$]{};
  \node[constant](beta_l)[below=of alpha_l, label=left:$\beta_\lambda$]{};
  \node[constant](nu)[right=of y, label=right:$\nu$]{};
  \node[constant](beta_g)[below=of gamma, label=below:$\beta_\gamma$]{};
  \node[constant](alpha_g)[left=of beta_g, label=below:$\alpha_\gamma$]{};
  \node[constant](beta_et)[right=of eta, label=right:$\beta_\eta$]{};
  \node[constant](alpha_et)[above=of beta_et, label=right:$\alpha_\eta$]{};
  
  \node[plate=\small{$t = 1:T$}, inner sep=1.3em, fit= (K) (H) (f) (lambda) (A) (y)]{};
  
  \path (K) edge [connect] (H)
  (H) edge [connect] (f)
  (lambda) edge [connect] (A)
  (A) edge [connect] (H)
  (f) edge [connect] (y)
  (w) edge [connect] (f)
  (b) edge [connect] (f)
  (gamma) edge [connect] (b)
  (eta) edge [connect] (w)
  (alpha_l) edge [connect] (lambda)
  (beta_l) edge [connect] (lambda)
  (alpha_g) edge [connect] (gamma)
  (beta_g) edge [connect] (gamma)
  (alpha_et) edge [connect] (eta)
  (beta_et) edge [connect] (eta)
  (nu) edge [connect] (y);

  \end{tikzpicture}
\caption{Directed graphical model for binary classification KBTL.}\label{fig:DGM_kbtl} 
\end{figure}

The DGM associated with the model is shown in Figure~\ref{fig:DGM_kbtl}; this highlights the variable dependences and the associated prior distributions. %
The distributional assumptions are \emph{briefly} summarised, for details, refer to~\cite{gonen2014kernelized}. %
The prior for the elements $\vec{A}_t[i,s]$ of the projection matrix are (zero mean) normally distributed, with variance $\vec{\Lambda}_t[i,s]^{-1}$; in turn, the prior over $\vec{\Lambda}_t[i,s]$ is Gamma distributed. %
As a result, the observations are normally distributed in the latent space, i.e.\ $\vec{H}_t[s,i]$. %
For the coupled classifier, the prior for the bias $b$ is assumed to be (zero mean) normally distributed, with variance $\gamma^{-1}$, such that $\gamma$ is Gamma distributed. %
Similarly, the weights $\vec{w}[s]$ are (zero mean) normally distributed, with variance $\vec{\eta}[s]^{-1}$, such that $\vec{\eta}[s]$ is Gamma distributed. %
This leads to normal distributions over the functional classifier $\vec{f}_t[i]$. %
The label predictive equations are given by $p(y^{(t)}_* \mid f^{(t)}_*)$, passing $f^{(t)}_*$ through a truncated Gaussian, parameterised by $\nu$ \cite{Gardner2020b}. %

The hyperparameters associated with these assumptions are shown in the DGM, Figure~\ref{fig:DGM_kbtl}. %
To infer the parameters of the model, approximate inference is required. Following \cite{gonen2014kernelized}, a variational inference scheme is used; this utilises a lower bound on the marginal likelihood, to infer an \emph{approximation}, denoted $q$, of the full joint distribution of the parameters $p(\vec{\Theta}, \vec{\Xi} \mid \left\{\vec{K}_t, \vec{y}_t \right\}_{t=1}^{T})$ of the model. To achieve this, the posterior distribution is factorised as follows,

\begin{align}
p\left(\vec{\Theta}, \vec{\Xi} \mid \left\{\vec{K}_t, \vec{y}_t \right\}_{t=1}^{T}\right) &\approx q(\vec{\Theta}, \vec{\Xi}) \nonumber\\
&= \prod_{t=1}^T\left(q(\vec{\Lambda}_t)q(\vec{A}_t)q(\vec{H}_t)\right)q(\gamma)q(\vec{\eta})q(b,\vec{w})\prod_{t=1}^T q(\vec{f}_t)
\end{align}

Each approximated factor is defined as in the full conditional distribution \cite{gonen2014kernelized}. %
The lower bound can be optimised with respect to each factor separately, while fixing the remaining factors (iterating until convergence). %

\subsubsection{Numerical + experimental example: Shear-building structures}
\begin{figure}[pt]
	\centering
	\begin{subfigure}{.25\textwidth}
	\centering
	\includegraphics[width=\linewidth]{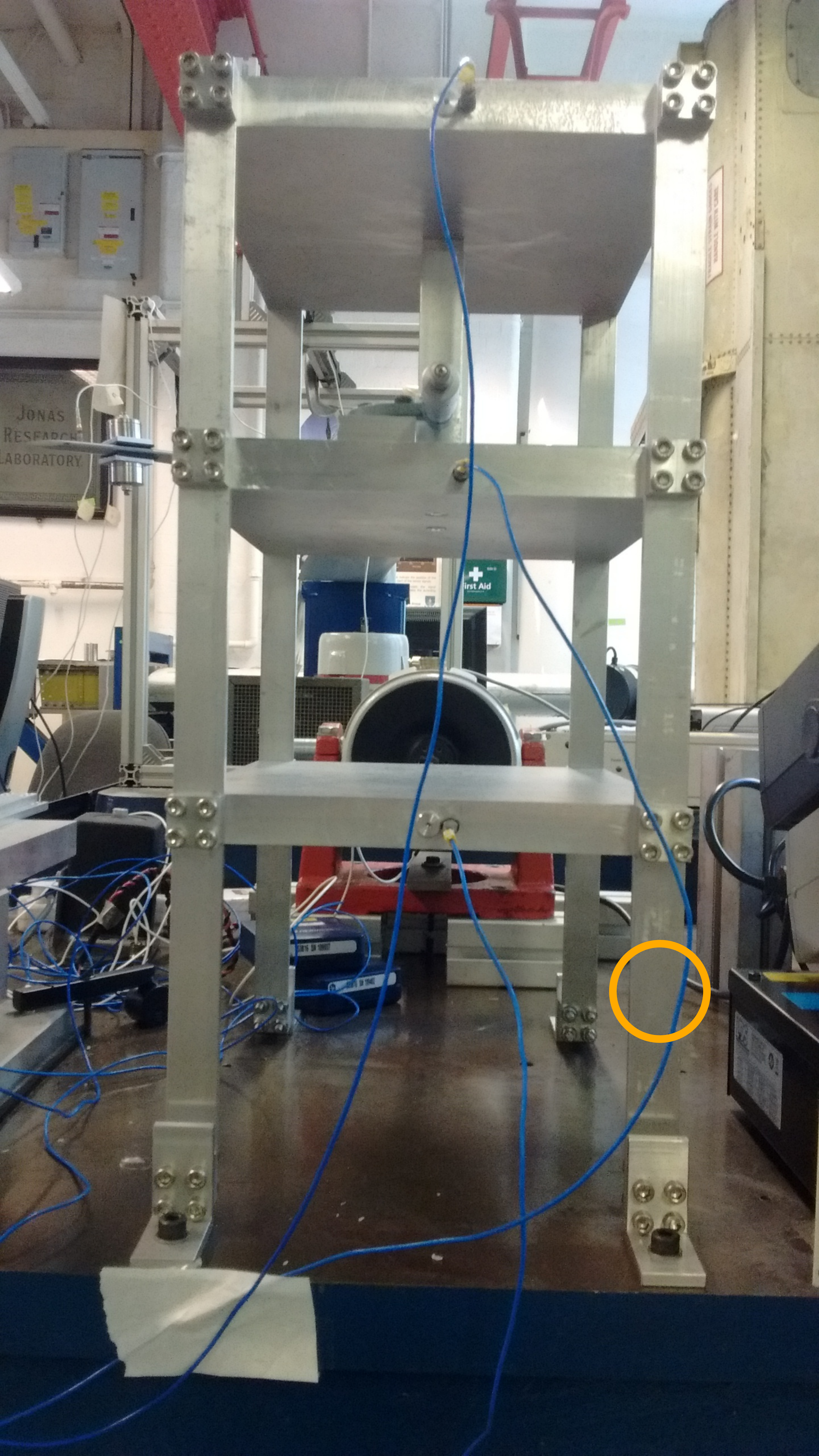}

	\caption{\label{a}}
	\label{fig:scehm_a}
	\end{subfigure}
	\begin{minipage}{.59\textwidth}
	\centering
	\resizebox{.85\linewidth}{!}{%
	\includegraphics[width=\linewidth]{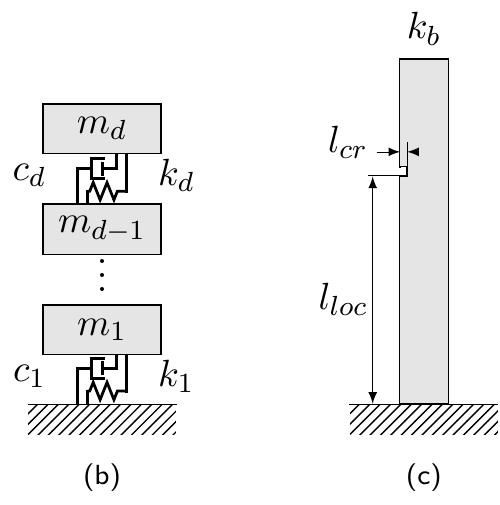}
	}
	\end{minipage}
	\caption{Shear structures: (a) test rig; (b) a nominal representation of the five simulated systems; (c) depicts the cantilever beam component where $\{k_i\}^d_{i=1} = 4k_b$. %
	}\label{fig:dofs}
\end{figure}

A numerical case study, supplemented with experimental data, is used for demonstration -- an extension of the work in \cite{imac_kbtl}. %
A population of six different shear-building structures is considered, five are simulated, and one is experimental. %
A domain and task are associated with each structure (such that $T=6$) -- the experimental rig and (simulated) lumped-mass models are shown in Figure \ref{fig:dofs}. %
For each structure (domain) there is a two-class classification problem (task), which is viewed as binary damage detection (normal or damaged). %

Each \textit{simulated} structure is represented by $d$ mass, stiffness and damping coefficients, i.e.\ $\{m_i, k_i, c_i\}^d_{i=1}$. %
The masses have length $l_m$, width $w_m$, thickness $t_m$, and density $\rho$. The stiffness elements are calculated from four cantilever beams in bending, $4k_b = 4(3EI/l_b^3)$, where $E$ is the elastic modulus, $I$ the second moment of area, and $l_b$ the length of the beam. %
The damping coefficients are specified rather than derived from a physical model. %
Damage is simulated via an open crack, using a reduction in $EI$ \cite{Christides1984}. %
For each structure, each observation is a random draw from a base distribution for $E$, $\rho$ and $c$. %
The properties of the five simulated structures are shown in Table \ref{tab:props}.

\begin{table}[h]
    \centering
    \caption{Properties of the five simulated structures. Degrees-of-freedom (DOF) are denoted $d$.}\label{tab:props}
    \resizebox{\linewidth}{!}{%
    \begin{tabular}{ccccccc}
        \hline
        \rotatebox{-90}{\textbf{Domain~}} & \rotatebox{-90}{DOF} & \rotatebox{-90}{\makecell{Beam \\ dim.}} & \rotatebox{-90}{\makecell{Mass \\ dim.}} & \rotatebox{-90}{\makecell{Elastic \\ mod.}} & \rotatebox{-90}{Density} & \rotatebox{-90}{\makecell{Damping \\ coeff.}} \\
        ($t$) & ($d_t$) & $\{l_b,\,w_b,\,t_b\}$ & $\{l_m,\,w_m,\,t_m\}$ & $E$ & $\rho$ & $c$ \\
        &  & $\mathrm{mm}$ & $\mathrm{mm}$ & $\mathrm{GPa}$ & $\mathrm{kg/m^3}$ & $\mathrm{Ns/m}$ \\
        \hline
        1 & 4 & $\{185, 25, 6.35\}$ & $\{350, 254, 25\}$ & $\gaussianDist{71}{1.0\times10^{-9}}$ & $\gaussianDist{2700}{10}$ & $\gammaDist{50}{0.1}$ \\
        2 & 8 & $\{200, 35, 6.25\}$ & $\{450, 322, 35\}$ & $\gaussianDist{70}{1.2\times10^{-9}}$ & $\gaussianDist{2800}{22}$ & $\gammaDist{8}{0.8}$ \\
        3 & 10 & $\{177, 45, 6.15\}$ & $\{340, 274, 45\}$ & $\gaussianDist{72}{1.3\times10^{-9}}$ & $\gaussianDist{2550}{25}$ & $\gammaDist{25}{0.2}$ \\
        4 & 3 & $\{193, 32, 5.55\}$ & $\{260, 265, 32\}$ & $\gaussianDist{75}{1.5\times10^{-9}}$ & $\gaussianDist{2600}{15}$ & $\gammaDist{20}{0.1}$ \\
        5 & 5 & $\{165, 46, 7.45\}$ & $\{420, 333, 46\}$ & $\gaussianDist{73}{1.4\times10^{-9}}$ & $\gaussianDist{2650}{20}$ & $\gammaDist{50}{0.1}$ \\
        \hline
    \end{tabular}
    }
\end{table}

The experimental structure is constructed from aluminium 6082, with dimensions nominally similar to those in Table \ref{tab:props}. Observational data (the first three natural frequencies) were collected via model testing, where an electrodynamic shaker applied up to 6553.6 Hz broadband white-noise excitation containing 16384 spectral lines (0.2 Hz resolution). Forcing was applied to the first storey, and three uni-axial accelerometers measured the response at all storeys. %
Damage was artificially introduced as a 50\% saw-cut to the-mid point of the front-right beam in Figure~\ref{fig:dofs}a.

In each domain, the damped natural frequencies act as features, such that ${\vec{X}_t[i,:] = \{\omega_{i}\}^d_{i=1}}$. Therefore, as each domain has different DOFs/dimensions, heterogeneous transfer is required. %
The label set is consistent across all domains, corresponding to normal or damaged, i.e $y_i \in \{-1,1\}$ respectively. %
The training and test data for each domain are summarised in Table \ref{tab:datapoints}. %
The training data have various degrees of class imbalance, to reflect scenarios where certain structures in SHM provide more information about a particular state.

\begin{table}[h]
\centering
\caption{Number of data for all domains (numerical and experimental*).}\label{tab:datapoints}
\begin{small}
\begin{tabular}{lcccc}
    \hline
    \textbf{Domain} & \multicolumn{2}{c}{\textbf{Training}} & \multicolumn{2}{c}{\textbf{Testing}} \\
    (t) & $y = -1$ & $y = +1$ & $y = -1$ & $y = +1$ \\
    \hline
    1 & 250 & 100 & 500 & 500 \\
    2 & 100 & 25 & 500 & 500 \\
    3 & 120 & 20 & 500 & 500 \\
    4 & 200 & 150 & 500 & 500 \\
    5 & 500 & 10 & 500 & 500 \\
    6* & 3 & 3 & 2 & 2 \\
    \hline
\end{tabular}
\end{small}
\end{table}

\begin{figure}[pt]
\centering
\includegraphics[width=.8\textwidth]{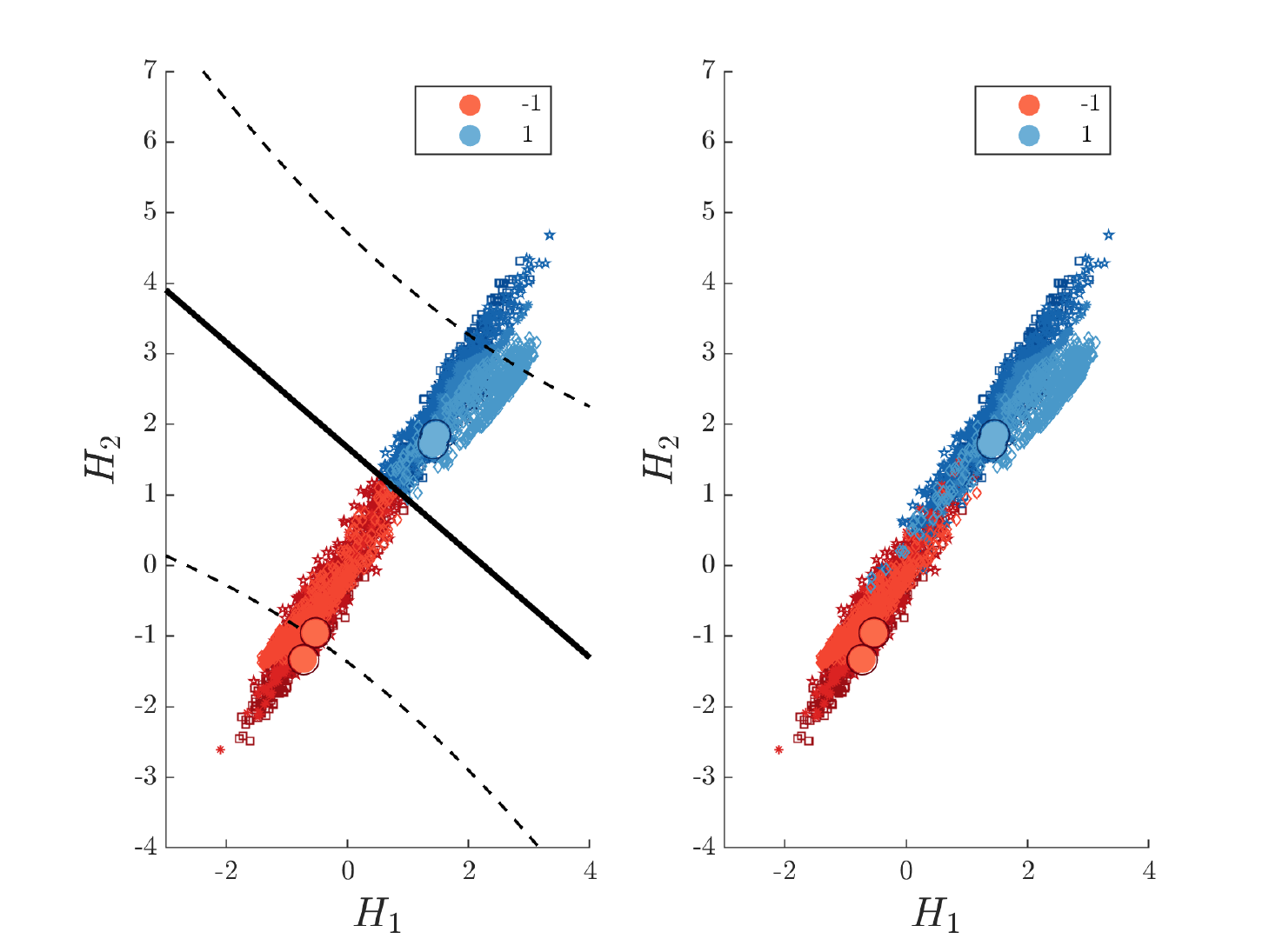}
\caption{The KBTL probabilistic decision boundary for the coupled classification model in the shared subspace. Markers $\{\times, \square, \star, *, \diamond, \triangle, \bullet\}$ correspond to tasks and domains $\{1,2,3,4,5,6\}$ respectively.}\label{fig:KBTL_hspace}
\end{figure}

Figure~\ref{fig:KBTL_hspace} shows the coupled binary classifier in the (expected) shared latent subspace for all the data $\left\{\vec{H}_t\right\}_{t=1}^T$. %
The observations associated with each of the six domains are distinguished via different markers. %
The left plot shows the test data and their predicted labels given $\vec{f}_t$, while the right plot shows the ground truth labels. %
KBTL has successfully embedded and projected data from different domains into a shared latent space ($R=2$), where the data can be categorised by a coupled discriminative classifier. %
It can also be seen that, due to class imbalance (weighted towards the undamaged class $-1$ for each structure), there is greater uncertainty in the damaged class ($+1$), leading to more significant scatter in the latent space. %

\begin{figure}[pt]
\centering
\includegraphics[width=.7\textwidth]{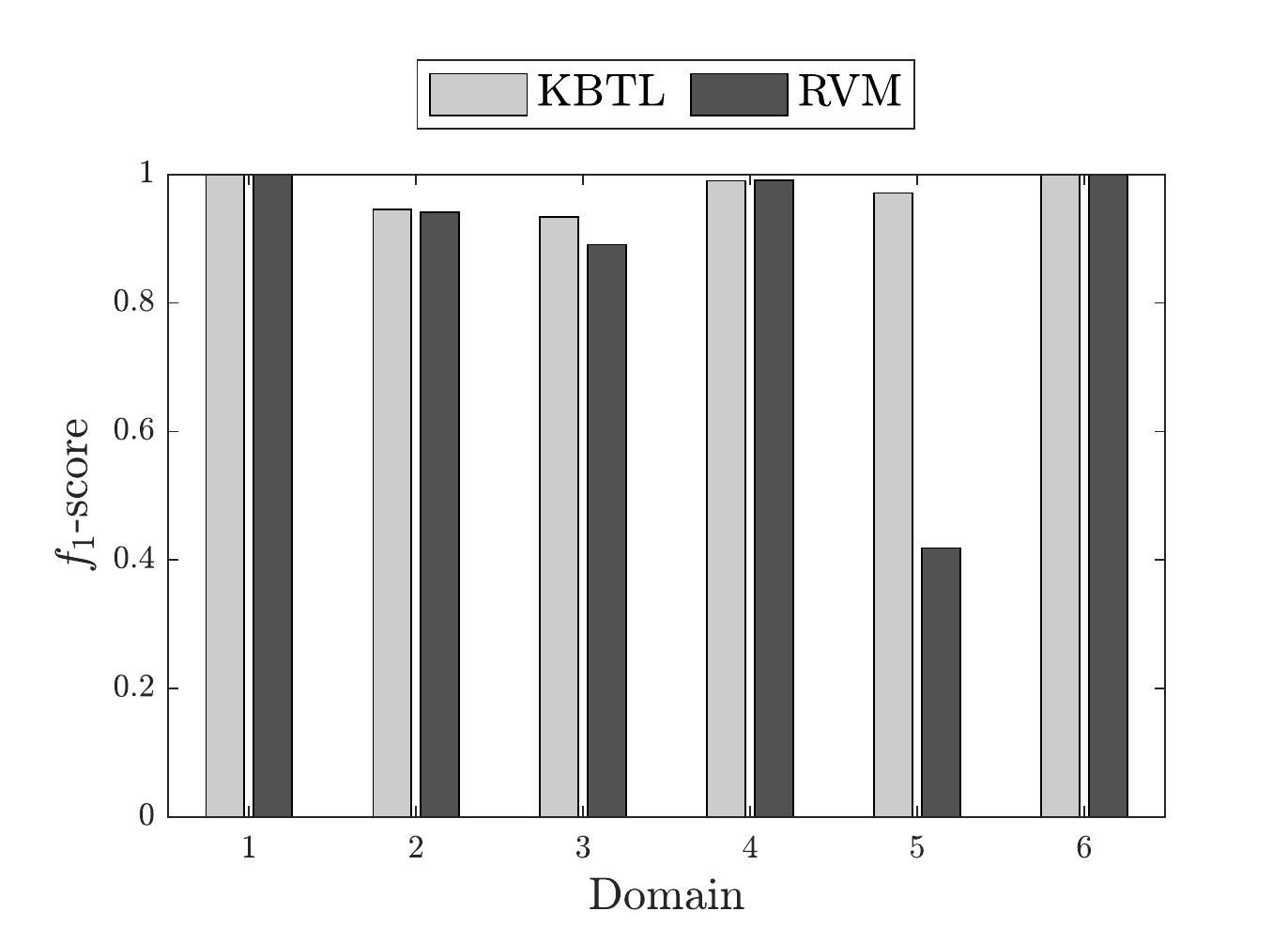}
\caption{KBTL classification performance, given an independent test set: $f_1$-scores across each domain compared to an RVM benchmark.}\label{fig:kbtl_f1}
\end{figure}

The classification results for each domain are presented in Figure~\ref{fig:kbtl_f1}. %
An observations is considered to belong to class $+1$ if $p(\vec{y}_{t}[*] = \,+1 \mid \vec{f}_t[*]) \geq 0.5$. %
KBTL is compared to a relevance vector machine (RVM) \cite{tipping2000relevance} as a benchmark -- learnt for each domain independently. %
It is acknowledged that the RVM differs in implementation; however, similarities make it useful for comparison as a standard (non multi-task) alternative to KBTL. %

Multi-task learning has accurately inferred a general model. %
For domains $\{1,2,3,5,6\}$, the SHM task is improved by considering the data from all structures in a shared latent space. %
In particular, extending the (effective) training data has improved the classification for domain 5. 
This is because there are few training data associated with the damage class for domain 5 (see Table~\ref{tab:datapoints}); therefore, considering damage data from similar structures (in the latent space) has proved beneficial. %
Interestingly, for domain four ($t=4$) there is a marginal \emph{decrease} in the classification performance. %
Like domain one, domain four has \textit{less} severe class imbalance, thus, it appears that the remaining domains (with severe class imbalance) have negatively impacted the score for this specific domain/task. %

These results highlight that the data from a group (or population) of \textit{similar} structures can be considered together, to increase the (effective) amount of training data \cite{PBSHMMSSP1,PBSHMMSSP2,PBSHMMSSP3}. %
This can lead to significant improvements in the predictive performance of SHM tools -- particularly those learnt from small sets of supervised data. %

\section{Conclusions}
Three new techniques for statistical inference with SHM signals have been collected and summarised (originally introduced in previous work), including %
partially-supervised learning (semi-supervised/active learning), Dirichlet process clustering, and multi-task learning. %
Primarily, each approach looks to address, from a different perspective, the issues of incomplete datasets and missing information, which lead to incomplete training-data. %
The algorithms consider that: a) label information (to describe what measurements represent) is likely to be incomplete; b) the available data \textit{a priori} will usually correspond to a \textit{subset} of the expected \textit{in situ} conditions only. %
Considering the importance of uncertainty quantification in SHM, probabilistic methods are suggested, which can be (intuitively) updated to account for missing information. %

The case study applications for each mode of inference highlight the potential advantages for SHM. %
Partially-supervised methods for active and semi-supervised learning were utilised to manage the cost system inspections (to label data), while considering the unlabelled instances, both offline and online. Dirichlet process clustering has been applied to streaming data, as an unsupervised method for automatic damage detection and classification. %
Finally multi-task learning was applied to model shared information between systems -- to extend the data available for training, this approach considers multiple (potentially incomplete) datasets associated with different tasks (structures).

\section{Data Availability}
Some or all data, models, or code that support the findings of this study are available from the corresponding author upon reasonable request.

\section{Acknowledgements}
The authors gratefully acknowledge the support of the UK Engineering and Physical Sciences Research Council (EPSRC) through grant references EP/R003645/1, EP/R004900/1, EP/S001565/1 and EP/R006768/1.

\appendix\label{section:references}

\bibliography{refs}

\begin{thebibliography}{}

\bibitem[\protect\citeauthoryear{}{Aldous}{1985}]{aldous1985exchangeability}
Aldous, D.~J. (1985).
\newblock ``Exchangeability and related topics.''\ {\em {\'E}cole d'{\'E}t{\'e}
  de Probabilit{\'e}s de Saint-Flour XIII—1983}, Springer,  1--198.

\bibitem[\protect\citeauthoryear{}{Barber}{2012}]{barber2012bayesian}
Barber, D. (2012).
\newblock {\em Bayesian Reasoning and Machine Learning}.
\newblock Cambridge University Press.

\bibitem[\protect\citeauthoryear{}{Blei et~al.\@}{2006}]{blei2006variational}
Blei, D.~M., Jordan, M.~I., et~al.\@ (2006).
\newblock ``Variational inference for {D}irichlet process mixtures.''\ {\em
  Bayesian Analysis}, 1(1), 121--143.

\bibitem[\protect\citeauthoryear{}{Bornn et~al.\@}{2009}]{bornn2009structural}
Bornn, L., Farrar, C.~R., Park, G., and Farinholt, K. (2009).
\newblock ``Structural health monitoring with autoregressive support vector
  machines.''\ {\em Journal of Vibration and Acoustics}, 131(2).

\bibitem[\protect\citeauthoryear{}{Bull et~al.\@}{2020a}]{PBSHMMSSP1}
Bull, L., Gardner, P., Gosliga, J., Dervilis, N., Papatheou, E., Maguire, A.,
  Campos, C., Rogers, T., Cross, E., and Worden, K. (2020a).
\newblock ``Foundations of population-based structural health monitoring,
  {P}art {I}: {H}omogeneous populations and forms.''\ {\em Preprint submitted
  to Mechanical Systems and Signal Processing}.

\bibitem[\protect\citeauthoryear{}{Bull}{2019}]{bull_2019thesis}
Bull, L.~A. (2019).
\newblock ``Towards probabilistic and partially-supervised structural health
  monitoring.''\ Ph.D. thesis, University of Sheffield, University of
  Sheffield.

\bibitem[\protect\citeauthoryear{}{Bull et~al.\@}{2019a}]{bull2019stsd}
Bull, L.~A., Manson, G., Worden, K., and Dervilis (2019a).
\newblock ``{Active learning approaches to structural health monitoring}.''\
  {\em Special Topics in Structural Dynamics, Volume 5}, N. Dervilis, ed.,
  Springer International Publishing,  157--159.

\bibitem[\protect\citeauthoryear{}{Bull et~al.\@}{2019b}]{onlineAL}
Bull, L.~A., Rogers, T.~J., Wickramarachchi, C., Cross, E.~J., Worden, K., and
  Dervilis, N. (2019b).
\newblock ``Probabilistic active learning: An online framework for structural
  health monitoring.''\ {\em Mechanical Systems and Signal Processing}, 134,
  106294.

\bibitem[\protect\citeauthoryear{}{Bull et~al.\@}{2019c}]{bull2019damage}
Bull, L.~A., Worden, K., and Dervilis, N. (2019c).
\newblock ``Damage classification using labelled and unlabelled
  measurements.''\ {\em Structural Health Monitoring 2019}.

\bibitem[\protect\citeauthoryear{}{Bull et~al.\@}{2020b}]{BULL2020106653}
Bull, L.~A., Worden, K., and Dervilis, N. (2020b).
\newblock ``Towards semi-supervised and probabilistic classification in
  structural health monitoring.''\ {\em Mechanical Systems and Signal
  Processing}, 140, 106653.

\bibitem[\protect\citeauthoryear{}{Bull et~al.\@}{2018}]{bull2018active}
Bull, L.~A., Worden, K., Manson, G., and Dervilis, N. (2018).
\newblock ``Active learning for semi-supervised structural health
  monitoring.''\ {\em Journal of Sound and Vibration}, 437, 373--388.

\bibitem[\protect\citeauthoryear{}{Bull
  et~al.\@}{2020c}]{bull2020investigating}
Bull, L.~A., Worden, K., Rogers, T.~J., Cross, E.~J., and Dervilis, N. (2020c).
\newblock ``Investigating engineering data by probabilistic measures.''\ {\em
  Special Topics in Structural Dynamics \& Experimental Techniques, Volume 5},
  Springer,  77--81.

\bibitem[\protect\citeauthoryear{}{Bull et~al.\@}{2019d}]{bull2019machiningAL}
Bull, L.~A., Worden, K., Rogers, T.~J., Wickramarachchi, C., Cross, E.~J.,
  McLeay, T., Leahy, W., and Dervilis, N. (2019d).
\newblock ``A probabilistic framework for online structural health monitoring:
  Active learning from machining data streams.''\ {\em Journal of Physics:
  Conference Series}, Vol. 1264, IOP Publishing,  012028.

\bibitem[\protect\citeauthoryear{}{Cappello
  et~al.\@}{2015}]{cappello2015mechanical}
Cappello, C., Bolognani, D., and Zonta, D. (2015).
\newblock ``Mechanical equivalent of logical inference from correlated
  uncertain information.''\ {\em Proc. of 7th International Conference on
  Structural Health Monitoring of Intelligent Infrastructure}.

\bibitem[\protect\citeauthoryear{}{Chakraborty
  et~al.\@}{2011}]{Chakraborty2011}
Chakraborty, D., Kovvali, N., Chakraborty, B., Papandreou-Suppappola, A., and
  Chattopadhyay, A. (2011).
\newblock ``{Structural damage detection with insufficient data using transfer
  learning techniques}.''\ {\em Sensors and Smart Structures Technologies for
  Civil, Mechanical, and Aerospace Systems},  798147.

\bibitem[\protect\citeauthoryear{}{Chapelle et~al.\@}{2006}]{SS}
Chapelle, O., Scholkopf, B., and Zien, A. (2006).
\newblock {\em {Semi-Supervised Learning}}.
\newblock MIT press.

\bibitem[\protect\citeauthoryear{}{Chatzi and
  Smyth}{2009}]{chatzi2009unscented}
Chatzi, E.~N. and Smyth, A.~W. (2009).
\newblock ``The unscented {K}alman filter and particle filter methods for
  nonlinear structural system identification with non-collocated heterogeneous
  sensing.''\ {\em Structural Control and Health Monitoring: The Official
  Journal of the International Association for Structural Control and
  Monitoring and of the European Association for the Control of Structures},
  16(1), 99--123.

\bibitem[\protect\citeauthoryear{}{{Chen} et~al.\@}{2013}]{chen2013}
{Chen}, S., {Cerda}, F., {Guo}, J., {Harley}, J.~B., {Shi}, Q., {Rizzo}, P.,
  {Bielak}, J., {Garrett}, J.~H., and {Kovacevic}, J. (2013).
\newblock ``Multiresolution classification with semi-supervised learning for
  indirect bridge structural health monitoring.''\ {\em 2013 IEEE International
  Conference on Acoustics, Speech and Signal Processing},  3412--3416\ (May).

\bibitem[\protect\citeauthoryear{}{{Chen} et~al.\@}{2014}]{chen2014}
{Chen}, S., {Cerda}, F., {Rizzo}, P., {Bielak}, J., {Garrett}, J.~H., and
  {Kovacevic}, J. (2014).
\newblock ``Semi-supervised multiresolution classification using adaptive graph
  filtering with application to indirect bridge structural health
  monitoring.''\ {\em IEEE Transactions on Signal Processing}, 62(11),
  2879--2893.

\bibitem[\protect\citeauthoryear{}{Christides and Barr}{1984}]{Christides1984}
Christides, S. and Barr, A. (1984).
\newblock ``One-dimensional theory of cracked bernoulli-euler beams.''\ {\em
  International Journal of Mechanical Sciences}, 26(11-12), 639--648.

\bibitem[\protect\citeauthoryear{}{Cozman et~al.\@}{2003}]{cozman2003semi}
Cozman, F.~G., Cohen, I., and Cirelo, M.~C. (2003).
\newblock ``Semi-supervised learning of mixture models.''\ {\em Proceedings of
  the 20th International Conference on Machine Learning (ICML-03)},  99--106.

\bibitem[\protect\citeauthoryear{}{Dasgupta}{2011}]{two_faces}
Dasgupta, S. (2011).
\newblock ``Two faces of active learning.''\ {\em Theoretical Computer
  Science}, 412(19), 1767--1781.

\bibitem[\protect\citeauthoryear{}{de~Roeck}{2003}]{SIMCES}
de~Roeck, G. (2003).
\newblock ``The state-of-the-art of damage detection by vibration monitoring:
  the {SIMCES} experience.''\ {\em Structural Control and Health Monitoring},
  10(2), 127--134.

\bibitem[\protect\citeauthoryear{}{Dempster
  et~al.\@}{1977}]{dempster1977maximum}
Dempster, A.~P., Laird, N.~M., and Rubin, D.~B. (1977).
\newblock ``Maximum likelihood from incomplete data via the {EM} algorithm.''\
  {\em Journal of the Royal Statistical Society: Series B (Methodological)},
  39(1), 1--22.

\bibitem[\protect\citeauthoryear{}{Dervilis et~al.\@}{2014}]{robust_og}
Dervilis, N., Cross, E., Barthorpe, R., and Worden, K. (2014).
\newblock ``Robust methods of inclusive outlier analysis for structural health
  monitoring.''\ {\em Journal of Sound and Vibration}, 333(20), 5181--5195.

\bibitem[\protect\citeauthoryear{}{Dorafshan et~al.\@}{2018}]{Dorafshan2018}
Dorafshan, S., Thomas, R.~J., and Maguire, M. (2018).
\newblock ``{Comparison of deep convolutional neural networks and edge
  detectors for image-based crack detection in concrete}.''\ {\em Construction
  and Building Materials}, 186, 1031--1045.

\bibitem[\protect\citeauthoryear{}{Farrar and Worden}{2012}]{SHM}
Farrar, C.~R. and Worden, K. (2012).
\newblock {\em Structural Health Monitoring: A Machine Learning Perspective}.
\newblock John Wiley \& Sons.

\bibitem[\protect\citeauthoryear{}{Flynn and Todd}{2010}]{flynn2010bayesian}
Flynn, E.~B. and Todd, M.~D. (2010).
\newblock ``A {B}ayesian approach to optimal sensor placement for structural
  health monitoring with application to active sensing.''\ {\em Mechanical
  Systems and Signal Processing}, 24(4), 891--903.

\bibitem[\protect\citeauthoryear{}{Gao and Mosalam}{2018}]{Gao2018}
Gao, Y. and Mosalam, K.~M. (2018).
\newblock ``{Deep transfer learning for image-based structural damage
  recognition}.''\ {\em Computer-Aided Civil and Infrastructure Engineering},
  33(9), 748--768.

\bibitem[\protect\citeauthoryear{}{Gardner et~al.\@}{2020a}]{imac_kbtl}
Gardner, P., Bull, L., Dervilis, N., and Worden, K. (2020a).
\newblock ``Kernelised {B}ayesian transfer learning for population-based
  structural health monitoring.''\ {\em Proceedings of IMAC XXXVIII, the
  38$^{th}$ International Modal Analysis Conference}, Springer.

\bibitem[\protect\citeauthoryear{}{Gardner et~al.\@}{2020b}]{PBSHMMSSP3}
Gardner, P., Bull, L., Gosliga, J., Dervilis, N., and Worden, K. (2020b).
\newblock ``Foundations of population-based structural health monitoring, part
  {III}: Heterogeneous populations -- mapping and transfer.''\ {\em Preprint
  submitted to Mechanical Systems and Signal Processing}.

\bibitem[\protect\citeauthoryear{}{Gardner et~al.\@}{2020c}]{Gardner2020b}
Gardner, P., Bull, L.~A., Dervilis, N., and Worden, K. (2020c).
\newblock ``A sparse {B}ayesian approach to heterogeneous transfer learning for
  population-based structural health monitoring.''\ {\em Submitted to
  Mechanical Systems and Signal Processing}.

\bibitem[\protect\citeauthoryear{}{Gardner et~al.\@}{2020d}]{GARDNER2020106550}
Gardner, P., Liu, X., and Worden, K. (2020d).
\newblock ``On the application of domain adaptation in structural health
  monitoring.''\ {\em Mechanical Systems and Signal Processing}, 138, 106550.

\bibitem[\protect\citeauthoryear{}{Gelman et~al.\@}{2013}]{gelman2013bayesian}
Gelman, A., Stern, H.~S., Carlin, J.~B., Dunson, D.~B., Vehtari, A., and Rubin,
  D.~B. (2013).
\newblock {\em Bayesian Data Analysis}.
\newblock Chapman and Hall/CRC.

\bibitem[\protect\citeauthoryear{}{G{\"o}nen and
  Margolin}{2014}]{gonen2014kernelized}
G{\"o}nen, M. and Margolin, A. (2014).
\newblock ``Kernelized {B}ayesian transfer learning.''\ {\em Twenty-Eighth AAAI
  Conference on Artificial Intelligence}.

\bibitem[\protect\citeauthoryear{}{Gosliga et~al.\@}{2020}]{PBSHMMSSP2}
Gosliga, J., Gardner, P., Bull, L., Dervilis, N., and Worden, K. (2020).
\newblock ``Foundations of population-based structural health monitoring, part
  {II}: Heterogeneous populations -- graphs, networks and communities.''\ {\em
  Preprint submitted to Mechanical Systems and Signal Processing}.

\bibitem[\protect\citeauthoryear{}{Huang et~al.\@}{2019}]{Yong2019}
Huang, Y., Beck, J.~L., and Li, H. (2019).
\newblock ``Multitask sparse bayesian learning with applications in structural
  health monitoring.''\ {\em Computer-Aided Civil and Infrastructure
  Engineering}, 34(9), 732--754.

\bibitem[\protect\citeauthoryear{}{Jang et~al.\@}{2019}]{Jang2019}
Jang, K., Kim, N., and An, Y. (2019).
\newblock ``{Deep learning-based autonomous concrete crack evaluation through
  hybrid image scanning}.''\ {\em Structural Health Monitoring},
  147592171882171.

\bibitem[\protect\citeauthoryear{}{Janssens et~al.\@}{2017}]{janssens2017deep}
Janssens, O., Van~de Walle, R., Loccufier, M., and Van~Hoecke, S. (2017).
\newblock ``Deep learning for infrared thermal image based machine health
  monitoring.''\ {\em IEEE/ASME Transactions on Mechatronics}, 23(1), 151--159.

\bibitem[\protect\citeauthoryear{}{Kremer et~al.\@}{2014}]{kremer_asvm}
Kremer, J., Steenstrup, K.~P., and Igel, C. (2014).
\newblock ``Active learning with support vector machines.''\ {\em Wiley
  Interdisciplinary Reviews: Data Mining and Knowledge Discovery}, 4(4),
  313--326.

\bibitem[\protect\citeauthoryear{}{MacKay}{2003}]{mackay2003information}
MacKay, D.~J. (2003).
\newblock {\em Information Theory, Inference and Learning Algorithms}.
\newblock Cambridge University Press.

\bibitem[\protect\citeauthoryear{}{Manson et~al.\@}{2003}]{partIII}
Manson, G., Worden, K., and Allman, D. (2003).
\newblock ``Experimental validation of a structural health monitoring
  methodology: Part {III}. damage location on an aircraft wing.''\ {\em Journal
  of Sound and Vibration}, 259(2), 365--385.

\bibitem[\protect\citeauthoryear{}{McCallumzy and
  Nigamy}{1998}]{mccallumzy1998employing}
McCallumzy, A.~K. and Nigamy, K. (1998).
\newblock ``Employing {EM} and pool-based active learning for text
  classification.''\ {\em Proc. International Conference on Machine Learning
  (ICML)}, Citeseer,  359--367.

\bibitem[\protect\citeauthoryear{}{Murphy}{2012}]{murphy}
Murphy, K.~P. (2012).
\newblock {\em Machine Learning: a Probabilistic Perspective}.
\newblock MIT press.

\bibitem[\protect\citeauthoryear{}{Neal}{2000}]{neal2000markov}
Neal, R.~M. (2000).
\newblock ``Markov chain sampling methods for {D}irichlet process mixture
  models.''\ {\em Journal of Computational and Graphical Statistics}, 9(2),
  249--265.

\bibitem[\protect\citeauthoryear{}{Nigam et~al.\@}{1998}]{nigam1998learning}
Nigam, K., McCallum, A., Thrun, S., and Mitchell, T. (1998).
\newblock ``Learning to classify text from labeled and unlabeled documents.''\
  {\em AAAI/IAAI}, 792, 6.

\bibitem[\protect\citeauthoryear{}{Ou et~al.\@}{2017}]{ou2017vibration}
Ou, Y., Chatzi, E.~N., Dertimanis, V.~K., and Spiridonakos, M.~D. (2017).
\newblock ``Vibration-based experimental damage detection of a small-scale wind
  turbine blade.''\ {\em Structural Health Monitoring}, 16(1), 79--96.

\bibitem[\protect\citeauthoryear{}{Pan and Yang}{2009}]{pan2009survey}
Pan, S.~J. and Yang, Q. (2009).
\newblock ``A survey on transfer learning.''\ {\em IEEE Transactions on
  Knowledge and Data Engineering}, 22(10), 1345--1359.

\bibitem[\protect\citeauthoryear{}{Papoulis}{1965}]{papoulis1965}
Papoulis, A. (1965).
\newblock {\em Probabilities, Random Variables, and Stochastic Processes}.
\newblock McGraw-Hill.

\bibitem[\protect\citeauthoryear{}{Peeters and de~Roeck}{2001}]{OGz24}
Peeters, B. and de~Roeck, G. (2001).
\newblock ``One-year monitoring of the {Z}24-bridge: environmental effects
  versus damage events.''\ {\em Earthquake Engineering \& Structural Dynamics},
  30(2), 149--171.

\bibitem[\protect\citeauthoryear{}{Rasmussen}{2000}]{rasmussen2000igmm}
Rasmussen, C.~E. (2000).
\newblock ``The infinite {Gaussian} mixture model.''\ {\em Advances in Neural
  Information Processing Systems},  554--560.

\bibitem[\protect\citeauthoryear{}{Rasmussen and
  Ghahramani}{2001}]{rasmussen2001occam}
Rasmussen, C.~E. and Ghahramani, Z. (2001).
\newblock ``Occam's razor.''\ {\em Advances in neural information processing
  systems},  294--300.

\bibitem[\protect\citeauthoryear{}{Rippengill et~al.\@}{2003}]{AE}
Rippengill, S., Worden, K., Holford, K.~M., and Pullin, R. (2003).
\newblock ``Automatic classification of acoustic emission patterns.''\ {\em
  Strain}, 39, 31--41.

\bibitem[\protect\citeauthoryear{}{Rogers et~al.\@}{2019}]{rogers2019}
Rogers, T.~J., Worden, K., Fuentes, R., Dervilis, N., Tygesen, U.~T., and
  Cross, E.~J. (2019).
\newblock ``A {B}ayesian non-parametric clustering approach for semi-supervised
  structural health monitoring.''\ {\em Mechanical Systems and Signal
  Processing}, 119, 100 -- 119.

\bibitem[\protect\citeauthoryear{}{Rousseeuw and Driessen}{1999}]{fastmcd}
Rousseeuw, P.~J. and Driessen, K.~V. (1999).
\newblock ``A fast algorithm for the minimum covariance determinant
  estimator.''\ {\em Technometrics}, 41(3), 212--223.

\bibitem[\protect\citeauthoryear{}{Schwenker and Trentin}{2014}]{Schwenker2014}
Schwenker, F. and Trentin, E. (2014).
\newblock ``Pattern classification and clustering: a review of partially
  supervised learning approaches.''\ {\em Pattern Recognition Letters}, 37(1),
  4--14.

\bibitem[\protect\citeauthoryear{}{Settles}{2012}]{settles2012active}
Settles, B. (2012).
\newblock ``Active learning.''\ {\em Synthesis Lectures on Artificial
  Intelligence and Machine Learning}, 6(1), 1--114.

\bibitem[\protect\citeauthoryear{}{Sohn et~al.\@}{2003}]{sohn2003review}
Sohn, H., Farrar, C.~R., Hemez, F.~M., Shunk, D.~D., Stinemates, D.~W., Nadler,
  B.~R., and Czarnecki, J.~J. (2003).
\newblock ``A review of structural health monitoring literature: 1996--2001.''\
  {\em Los Alamos National Laboratory, USA}.

\bibitem[\protect\citeauthoryear{}{Tipping}{2000}]{tipping2000relevance}
Tipping, M.~E. (2000).
\newblock ``The relevance vector machine.''\ {\em Advances in Neural
  Information Processing Systems},  652--658.

\bibitem[\protect\citeauthoryear{}{Vanik et~al.\@}{2000}]{vanik2000bayesian}
Vanik, M.~W., Beck, J.~L., and Au, S. (2000).
\newblock ``Bayesian probabilistic approach to structural health monitoring.''\
  {\em Journal of Engineering Mechanics}, 126(7), 738--745.

\bibitem[\protect\citeauthoryear{}{Vlachos
  et~al.\@}{2009}]{vlachos2009unsupervised}
Vlachos, A., Korhonen, A., and Ghahramani, Z. (2009).
\newblock ``Unsupervised and constrained {D}irichlet process mixture models for
  verb clustering.''\ {\em Proceedings of the Workshop on Geometrical Models of
  Natural Language Semantics}, Association for Computational Linguistics,
  74--82.

\bibitem[\protect\citeauthoryear{}{Wan and Ni}{2019}]{Ping2019}
Wan, H. and Ni, Y. (2019).
\newblock ``Bayesian multi-task learning methodology for reconstruction of
  structural health monitoring data.''\ {\em Structural Health Monitoring}, 18,
  1282--1309.

\bibitem[\protect\citeauthoryear{}{Wang et~al.\@}{2017}]{wang_density}
Wang, M., Min, F., Zhang, Z.-H., and Wu, Y.-X. (2017).
\newblock ``Active learning through density clustering.''\ {\em Expert Systems
  with Applications}, 85, 305--317.

\bibitem[\protect\citeauthoryear{}{Worden and
  Manson}{2006}]{worden2006application}
Worden, K. and Manson, G. (2006).
\newblock ``The application of machine learning to structural health
  monitoring.''\ {\em Philosophical Transactions of the Royal Society A:
  Mathematical, Physical and Engineering Sciences}, 365(1851), 515--537.

\bibitem[\protect\citeauthoryear{}{Worden et~al.\@}{2008}]{genetic}
Worden, K., Manson, G., Hilson, G., and Pierce, S. (2008).
\newblock ``Genetic optimisation of a neural damage locator.''\ {\em Journal of
  Sound and Vibration}, 309(3), 529--544.

\bibitem[\protect\citeauthoryear{}{Ye et~al.\@}{2017}]{Ye2017}
Ye, J., Kobayashi, T., Tsuda, H., and Murakawa, M. (2017).
\newblock ``{Robust hammering echo analysis for concrete assessment with
  transfer learning}.''\ {\em Proceedings of the The 11th International
  Workshop on Structural Health Monitoring},  943--949.

\bibitem[\protect\citeauthoryear{}{Zhang and Yang}{2018}]{zhang2018overview}
Zhang, Y. and Yang, Q. (2018).
\newblock ``An overview of multi-task learning.''\ {\em National Science
  Review}, 5(1), 30--43.

\bibitem[\protect\citeauthoryear{}{Zhao et~al.\@}{2019}]{zhao2019deep}
Zhao, R., Yan, R., Chen, Z., Mao, K., Wang, P., and Gao, R.~X. (2019).
\newblock ``Deep learning and its applications to machine health monitoring.''\
  {\em Mechanical Systems and Signal Processing}, 115, 213--237.

\bibitem[\protect\citeauthoryear{}{Zhu}{2005}]{zhu2005semi}
Zhu, X.~J. (2005).
\newblock ``Semi-supervised learning literature survey.''\ {\em Report no.},
  University of Wisconsin-Madison Department of Computer Sciences.

\bibitem[\protect\citeauthoryear{}{Zonta et~al.\@}{2014}]{zonta2014value}
Zonta, D., Glisic, B., and Adriaenssens, S. (2014).
\newblock ``Value of information: impact of monitoring on decision-making.''\
  {\em Structural Control and Health Monitoring}, 21(7), 1043--1056.

\end{thebibliography}

\end{document}